\documentclass[10pt,journal,compsoc]{IEEEtran}
\makeatletter
\long\def\@makecaption#1#2{%
  \vskip\abovecaptionskip
  \sbox\@tempboxa{#1: #2}%
  \ifdim \wd\@tempboxa >\hsize
    #1: #2\par
  \else
    \global \@minipagefalse
    \hb@xt@\hsize{\box\@tempboxa\hfil}%
  \fi
  \vskip\belowcaptionskip}
\makeatother

\ifCLASSOPTIONcompsoc
  \usepackage[nocompress]{cite}
\else
  \usepackage{cite}
\fi
\usepackage{times}
\usepackage{amsmath}
\usepackage{amssymb}

\ifCLASSINFOpdf
  \usepackage[pdftex]{graphicx}
  \graphicspath{{../pdf/}{../jpeg/}}
  \DeclareGraphicsExtensions{.pdf,.jpeg,.png}
\else
\fi
\begin{document}
%
\title{The Devil is in the Details: Delving into Unbiased \\ Data Processing for Human Pose Estimation}
%
%
%
%

\author{Junjie~Huang,
        Zheng~Zhu,
        Feng~Guo,
        Guan~Huang,
        and~Dalong~Du
}

%
%

\markboth{}%
{Shell \MakeLowercase{\textit{et al.}}: Bare Demo of IEEEtran.cls for Computer Society Journals}
%



\IEEEtitleabstractindextext{%
\begin{abstract}
Being a fundamental component in training and inference, data processing has not been systematically considered in human pose estimation community, to the best of our knowledge. In this paper, we focus on this problem and find that the devil of human pose estimation evolution is in the biased data processing. Specifically, by investigating the standard data processing in state-of-the-art approaches mainly including coordinate system transformation and keypoint format transformation (i.e., encoding and decoding), we find that the results obtained by common flipping strategy are unaligned with the original ones in inference. Moreover, there is a statistical error in some keypoint format transformation methods. Two problems couple together, significantly degrade the pose estimation performance and thus lay a trap for the research community. This trap has given bone to many suboptimal remedies, which are always unreported, confusing but influential. By causing failure in reproduction and unfair in comparison, the unreported remedies seriously impedes the technological development. To tackle this dilemma from the source, we propose Unbiased Data Processing (UDP) consist of two technique aspect for the two aforementioned problems respectively (i.e., unbiased coordinate system transformation and unbiased keypoint format transformation). Base on UDP, we wipe out the trap by giving out a deep insight of the existing biased data processing pipeline, whose origin, effects and some confusing remedies are thoroughly studied. Besides, as a model-agnostic approach and a superior solution, UDP successfully pushes the performance boundary of human pose estimation. For example on COCO \textit{test-dev} set, UDP promotes top-down method HRNet-W32-256$\times$192 by 1.7 AP (73.5 to 75.2) for free and promotes bottom-up methods HRNet-W32-512$\times$512 by 2.7 AP with an acceleration of 6.1 times. The HRNet-W48-384$\times$288 equipped with UDP achieves 76.5 AP and sets a new state-of-the-art for human pose estimation. As a meaningful milestone for pursuing high performance human pose estimation, UDP has been the key base of the winner in 2020 COCO Keypoint Detection Challenge. The code is public available for reference.
\end{abstract}

}

\maketitle

\IEEEdisplaynontitleabstractindextext

%
\IEEEpeerreviewmaketitle

\section{Introduction}
\renewcommand{\thefootnote}{}
\footnotetext{J. Huang, F. Guo, G. Huang and D. Du are with XForwardAI Technology Co.,Ltd, Beijing, China. E-mail: junjie.huang@ieee.org, \{feng.guo, guan.huang, dalong.du\}@xforwardai.com}
\footnotetext{Z. Zhu is with Tsinghua University, Beijing, China. E-mail: zhengzhu@ieee.org}

2D Human pose estimation has been extensively studied in computer vision literature and serves many complicated downstream visual understanding tasks such as 3D human pose estimation \cite{human36m,ci2020locally,luvizon2020multi,3d8611195,R3D,MonoCap}, human phasing \cite{liang2018look,HumanParsing}, health care \cite{VBE,chen2020fall,chen2018patient}, video surveillance \cite{li2019state,zhang2019exploiting,andriluka2018posetrack,Girdhar_2018_CVPR} and action recognition \cite{carreira2017quo, zhu2019action, zhu2019convolutional,luvizon2020multi}. In this paper, we pay attention to the \textit{data processing} aspect, considering it as a fundamental component. All visual recognition tasks are born with data processing, and in general share data processing methodology with each other like data augmentation and transformation between different coordinate systems. However, when compared with other tasks like classification \cite{ImageNet}, object detection \cite{COCO} and semantic segmentation \cite{mottaghi2014role,cordts2016cityscapes}, the performance of human pose estimation algorithms is much more sensitive to the methods used in data processing on account of the evaluation principle. In the evaluation of human pose estimation, the metrics are calculated based on the positional offset between ground truth annotations and predicted results \cite{COCO,MPII}, where small disturbance caused by data processing will affect the performance of pose estimators by a large margin.


\begin{figure}[t]
	\setlength{\abovecaptionskip}{0.cm}
    \begin{center}
        \includegraphics[width=0.95\hsize]{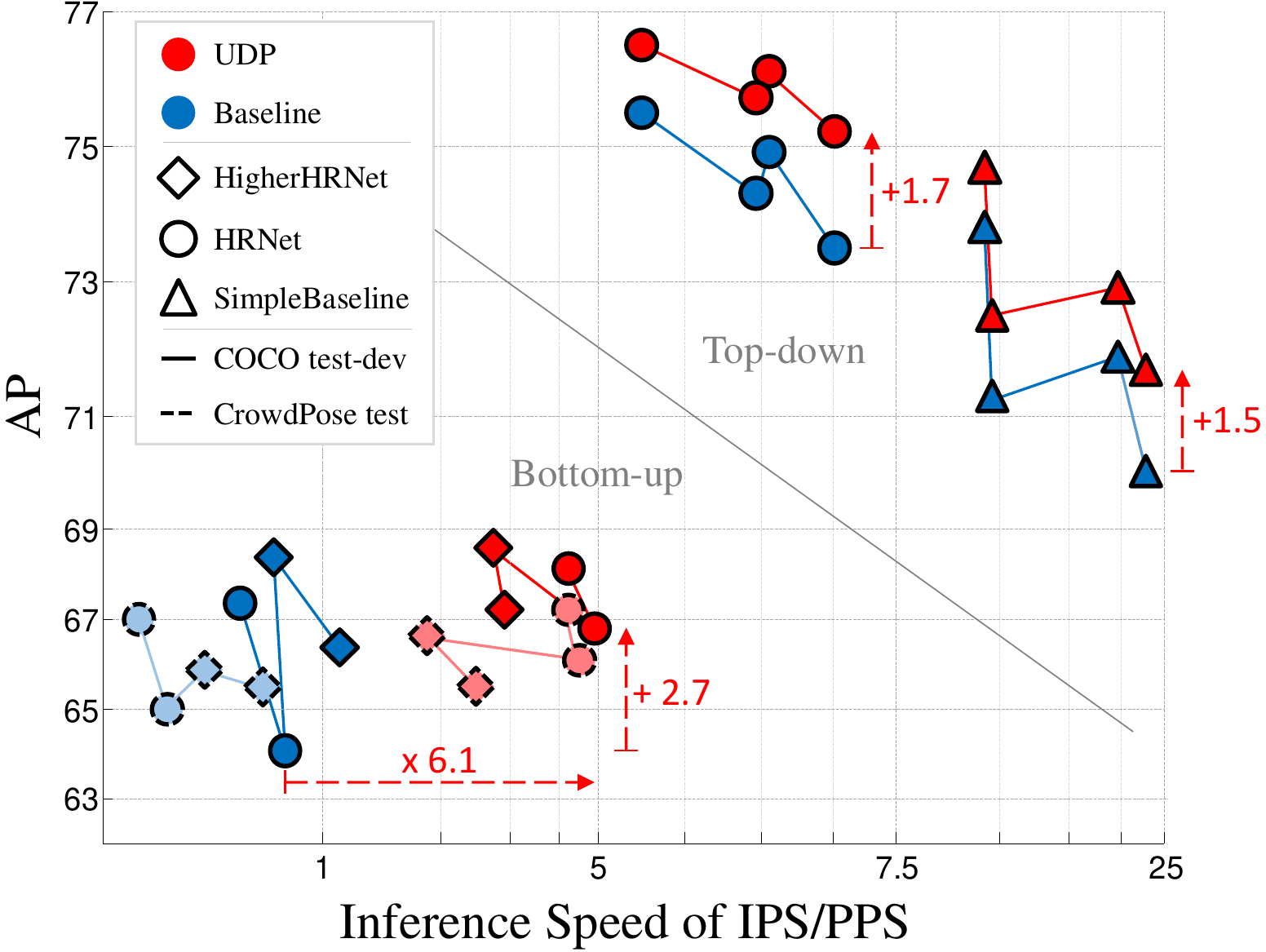}
    \end{center}
   \caption{The improvement of performance on COCO \textit{test-dev} set and CrowdPose \textit{test} set when the proposed Unbiased Data Processing (UDP) is applied to the state-of-the-art methods. At no cost, UDP improves the APs of top-down methods SimpleBaseline \cite{SBNet} and HRNet \cite{HRNet} by a considerable margin. In bottom-up paradigm, UDP offers both accuracy improvement and inference acceleration to HRNet \cite{HRNet} and HigherHRNet \cite{Higher}.}
    \label{fig:mAP-gflops}
\end{figure}

Although it is of significant, to the best of our knowledge, data processing has not been systematically considered in human pose estimation community. When this topic is addressed, we find that the widely used data processing pipelines in most state-of-the-art human pose estimation systems \cite{CPN,SBNet,MSPN,HRNet,Higher,DARK} are defective. The chief causes are two common problems: i) When flipping testing strategy is adopted, the results from flipped image are unaligned with those from the origin image. The bias derives from utilizing \textit{pixel} for measuring the size of images when performing coordinate system transformation in resizing operation. ii) Defective keypoint format transformation (i.e., encoding and decoding) methods would lead to extra precision degradation. The two problems accumulatively degrade the performance of human pose estimators, lay a trap for the research community and subsequently has given born to many suboptimal remedies. The empirical remedies are always unreported but with huge impact on the performance like direct compensation in post processing \cite{CPN,SBNet,HRNet,MSPN,DARK,Higher}, while the others are reported but at tremendous cost of latency like using higher network output resolution in HigherHRNet \cite{Higher}. It is worth noting that, by causing failure in reproduction and unfair in comparison, the unreported remedies will obstruct the development of the human pose estimation technologies.

In this paper, we offer a reasonable and free access to thoroughly solving the two aforementioned problems by proposing Unbiased Data Processing (UDP) system. Corresponding to the two aforementioned problems, UDP consists of two technical aspects: the unbiased coordinate system transformations and the  unbiased keypoint format transformations. Aiming at the unbiased coordinate system transformations, we firstly propose to follow the principle of defining and analyzing this problem in continuous space. Then the concept of coordinate system transformation is defined based on this principle and the targets of unbias in this sub problem are introduced. Subsequently, the coordinate system transformations in different elementary operations (e.g., cropping, resizing, rotating and flipping) are formally designed, which finally compose the common coordinate system transformations used in training and testing process. With mathematical reasoning, we verify the unbiased property of the designed coordinate system transformation pipeline, and subsequently offer a deep insight of the existing biased coordinate system transformation pipeline, whose origin, effects and some confusing remedies are thoroughly studied. Analogously, the concept of unbiased keypoint format transformation is proposed, two unbiased keypoint format transformation methods are introduced and a typical biased example is analyzed thoroughly. As a result with UDP, the aforementioned trap can be remove and a higher as well as more reliable baseline can be achieved.

To showcase the effectiveness of the proposed method, we perform comprehensive experiments on the COCO Keypoint Detection benchmarks \cite{COCO}. As a model-agnostic approach and a superior solution, UDP successfully pushes the performance boundary of the human pose estimation problem as illustrated in Figure~\ref{fig:mAP-gflops}. Specifically, UDP boosts the performance of the methods in top-down paradigm without any extra latency. For example, UDP promotes the SimpleBaseline \cite{SBNet} by 1.5 AP (70.2 to 71.7) and 1.0 AP (71.9 to 72.9) within ResNet50-256$\times$192 and ResNet152-256$\times$192 configurations, respectively. For HRNet \cite{HRNet} within W32-256$\times$192 and W48-256$\times$192 configurations, UDP obtains gains by 1.7 AP (73.5 to 75.2) and 1.4 AP (74.3 to 75.7), respectively. The HRNet-W48-384$\times$288 equipped with UDP achieves 76.5 AP (1.0 improvement) and sets a new state-of-the-art for top-down human pose estimation. Besides, in bottom-up paradigm, UDP simultaneously offers both accuracy improvement and latency reducing on the baselines. For HRNet-W32-512$\times$512 configuration in HigherHRNet \cite{Higher}, UDP promotes its performance by 2.7 AP, and at the same time, offers an acceleration by 6.1 times. For HigherHRNet-W32-512$\times$512 configuration, the promotion and acceleration are +0.8 AP and 2.6 times respectively. In addition, we also perform experiments on extra dataset CrowdPose \cite{Crowdpose} to verify the generalization ability of UDP among different data distributions. Experimental results show that the performance of UDP in this dataset is in line with that on COCO dataset. Finally to verify the statement in methodology analysis, we measure the contribution of each element in UDP and the effect of the existing remedies in relative works with exhaustive ablation study. Based on the experiment results, we call for attention on the data processing aspect when designing or evaluating the future works. The code is public available for reference\footnote{https://github.com/HuangJunJie2017/UDP-Pose}.

This paper is built upon our conference paper \cite{UDP} and significantly extended in several aspects. First, we rearrange the methodology section for methodical stating, and explain it with more specific background introduction and more detailed mathematical reasoning. Second, we extend the coverage of UDP by applying it to methods in bottom-up paradigm and make great discovery and promotion on state-of-the-art method HigherHRNet \cite{Higher}. Third, we use extra dataset CrowdPose \cite{Crowdpose} to verify the generalization ability of UDP. In COCO and LVIS 2020 competitions\footnote{https://cocodataset.org/workshop/coco-lvis-eccv-2020.html}, UDP serves as the baseline for the winner UDP++ \cite{UDP++}, which marks this work as a meaningful milestone for pursuing high performance human pose estimation.

\section{Related Work}
\label{sec:RW}
In recent years, research community has witnessed a significant advance from single person \cite{MPII, IEF,PS,ConvNetPOSE,DPM,DeepPose,tompson2014joint,CPM,Hourglass,fppose} to multi-person pose estimation \cite{jin2020whole,COCO,iqbal2016multi,DeepCut,DeeperCut,OpenPose,G-RMI,nie2019single,CPN,HRNet,AssociativeEmbedding,OccNet,Higher}, where the latter can be generally categorized into bottom-up \cite{DeepCut,DeeperCut,OpenPose,AssociativeEmbedding,PersonLab,Higher} and top-down  \cite{G-RMI,CPN,MSPN,CFA,Mask-RCNN,OccNet,SBNet,HRNet} approaches.

\textbf{Bottom-up methods} start by detecting identity-free joints for all the persons in an input image and then group them into person instances. In this paradigm, both cost and efficient are considered, both the identity-free joint detection and grouping strategy are the main concerns. OpenPose \cite{OpenPose} builds a model that contains two branches to predict keypoint heatmaps and pairwise relationships (part affinity fields) between them, where the latter acts as the main cue in grouping process. MultiPoseNet \cite{MultiPoseNet} simultaneously achieves human detection and pose estimation, and proposes PRN to group the keypoints by the bounding box of each people. Aiming at resolving the human pose estimation problem in crowd sense, Li et al. \cite{Crowdpose} design a new model by combining joint-candidate single person pose estimation and global maximum joints association. Simultaneously, a new dataset named CrowdPose is collected specific for performance evaluation in crowd senses. Newell et al. \cite{AssociativeEmbedding} use one network for both heatmap prediction and embedding study. Grouping is done by utilizing association embedding, which assigns each keypoint with a tag and groups keypoints based on the L2 distance between tag vectors. As a follower, Chen et al. \cite{Higher} replace the hourglass style networks in \cite{AssociativeEmbedding} with the proposed HigherHRNet. By using higher output resolution, HigherHRNet improves the precision of the predictions by a large margin.

\textbf{Top-down methods} achieve multi-person pose estimation by the two-stages process, including obtaining person bounding boxes through a person detector like Faster R-CNN \cite{ren2016faster} and predicting keypoint locations within these boxes. As single person pose estimation is performed with fixed scale patches, most state-of-the-art performances on multi-person popular benchmarks COCO \cite{COCO} are achieved by top-down methods \cite{CPN,MSPN,UDP}. Existing works with this paradigm pay more attention to the designing of network structure. Chen et al. \cite{SCN} propose Structure-aware Convolutional Network trained with Generative Adversarial Networks for human pose structure exploiting. Following ShuffleNet \cite{zhang2018shufflenet} and SENet \cite{hu2018squeeze}, Su et al. \cite{su2019multi} propose Channel Shuffle Module (CSM) and Spatial, Channel-wise Attention Residual Bottleneck (SCARB) specific for human pose estimation problem. CPN \cite{CPN} and MSPN \cite{MSPN} are the leading methods on COCO keypoint challenge, adopting cascade network to refine the keypoints prediction. SimpleBasline \cite{SBNet} adds a few deconvolutional layers to enlarge the resolution of output features. Thought simple, it has a competitive performance among existing works. HRNet \cite{HRNet} maintains high-resolution representations through the whole process, achieving state-of-the-art performance on public datasets. Mask R-CNN \cite{Mask-RCNN} builds an end-to-end framework and achieves a good balance between performance and inference speed.

\textbf{Data processing} in human pose estimation mainly includes \textit{coordinate system transformation} and \textit{keypoint format transformation}. \textbf{Coordinate system transformation} means transforming the data (i.e., keypoint coordinates and image matrixes) between different coordinate systems when some operations are conducted like cropping, resizing, rotating and flipping. During this process, most state-of-the-art methods \cite{CPN,SBNet,MSPN,HRNet,Higher} use \textit{pixel} to measure the size of images when performing resizing operation, leading to unaligned results when using flipping strategy in inference. This bias degrades the accuracy by a large margin, lays a trap for research community and has given bone to some suboptimal remedies. The remedies are all empirical and always unreported. For example, without any explanation, SimpleBaseline \cite{SBNet} HRNet \cite{HRNet} and Darkpose \cite{DARK} empirically shift the result from flipped image by 1 pixel in network output coordinate system to suppress the predicting error. CPN \cite{CPN} and MSPN \cite{MSPN} achieve similar effect by shifting the average result by 2 pixels in network input coordinate system. HigherHRNet \cite{Higher} proposes to use higher network output resolution and conducts the experiment with some unreported compensation for large superiority on the baseline. These remedies are effective and appealing, but being the recipe for disaster as they hinder the development of technology by causing failure in reproduction and unfair in comparison. In this paper, we propose unbiased coordinate system transformation to thoroughly solve this problem, which will not only boost the performance of the existing methods but also provide a more reliable baseline for future works. \textbf{Keypoint format transformation} (i.e., encoding and decoding) commonly denotes the transformation between joint coordinates and heatmaps, which is firstly proposed in \cite{tompson2014joint} and has been widely used in state-of-the-art methods \cite{Mask-RCNN,Higher,CPN,SBNet,MSPN,HRNet}. In training process, it encodes the annotated keypoint coordinate into a heatmap with Gaussian distribution. And in testing process, it decodes the network predicted heatmap back into keypoint coordinate. This pipeline shows superior performance when compared with directly predicting the keypoint coordinates \cite{sun2018integral}, but is still imperfect on account of its defective design and inherent precision degradation. The combined classification and regression format based encoding-decoding paradigm in \cite{G-RMI} provides an mathematically error-free entrance to further promote the prediction accuracy. Analogously, Darkpose \cite{DARK} achieve unbiased keypoint format transformation by proposing a distribution-aware decoding method to match the encoding method use in \cite{Higher}. In this paper, we will introduce these two unbiased keypoint format transformation paradigm, verify their unbias property and show their superiority on the baseline.

\section{Unbiased Data Processing for Human Pose Estimation}
\label{sec:MT}
In human pose estimation, data processing involves the transformation between different coordinate system and the transformation between different keypoint format. In the following, we will give details introduction of our unbiased data processing method in these two aspects respectively (i.e., \textit{unbiased coordinate system transformation} and \textit{unbiased keypoint format transformation}).

\subsection{Unbiased Coordinate System Transformation}
As it is new to this topic and quite ambiguous to community, for clarified and reasonable statement, the concept of unbiased coordinate system transformation is constructed from the base. We firstly propose \textit{the unified definition of data in continuous space}. Then based on this definition, the concept of \textit{coordinate system transformation} and \textit{the targets of unbias} are introduced. We design the coordinate system transformation in some elementary operations (i.e., cropping, resizing, rotating and flipping) before we construct the common composite transformations between the coordinate systems involved in human pose estimation problem(i.e., source image coordinate system, network input coordinate system and network output coordinate system). Subsequently, we verify the unbias properties of the designed coordinate system transformation pipeline with mathematical reasoning. And at last to showcase how the defective coordinate system affect the research community, some biased data processing methods are analyzed, and the theory behind some reported techniques and unreported tricks used in state-of-the-arts are thoroughly studied.

\label{sec:data_process}
\subsubsection{An Unified Definition of Data in Continuous Space.}
The image matrixes and the target keypoint coordinates are the main data involved in human pose estimation problem. The images are stored and processed in a discrete format, but the keypoint coordinates are defined, processed and evaluated in continuous spaces. To avoid precision degradation in the coordinate system transformation pipeline, an unified paradigm is required for uniformly analyzing and dealing with different data in the coordinate system transformation problems.

\begin{figure}[h]
    \centering
    \includegraphics[width=0.8\hsize]{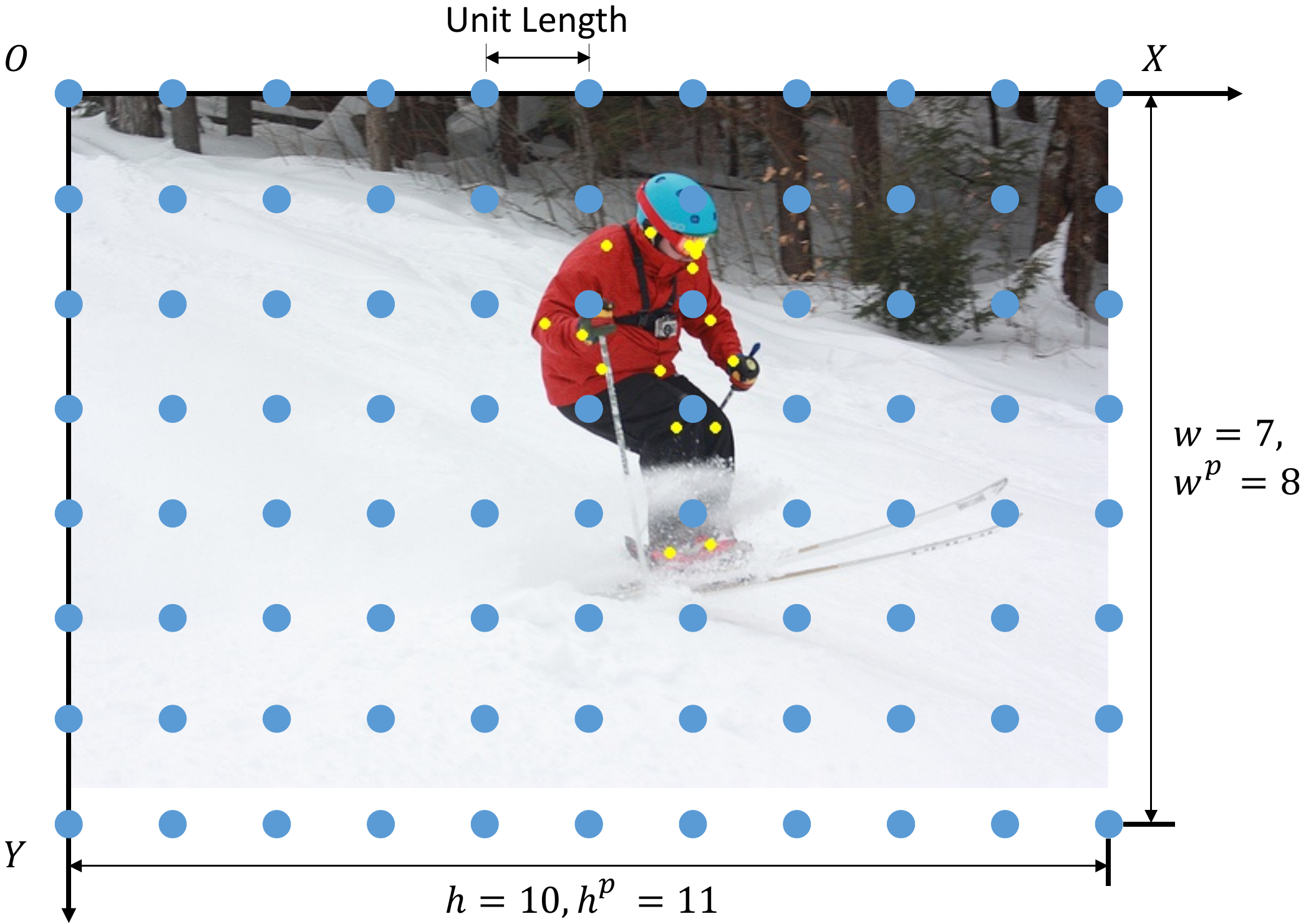}
    \caption{Illustration of analyzing the coordinate system transformation problem in continuous space. $O\text{-}XY$ denotes the coordinate system. An image matrix (the set of blue points) is regarded as a sampling result of the continuous image plane.}
    \label{fig:analysis_in_continuous_space}
\end{figure}

To this end, we assume that there is a continuous image plane and consider each image matrix as a discrete sampling result on it, where each pixel in an image matrix is a specific sample point. Formally, in line with the definition of target keypoint coordinates in COCO dataset\cite{COCO}, we define the coordinate system $O\text{-}XY$ as illustrated in Figure~\ref{fig:analysis_in_continuous_space} to describe the continuous image planes. The origin of the coordinate system is located at the most top-left pixel, the $O\text{-}X$ direction is from left to right and the $O\text{-}Y$ direction is from top to down. Besides, the distance between adjacent pixels is assumed to be equivalent and is defined as the unit length of the coordinate system. Then we have an image matrix as a sampling result of the image plane $\textbf{I}$, which is denoted as $\{\textbf{I}(\textbf{p})=(r,g,b)|\textbf{p}=(x,y), x\in\{0,1,2...w\}, y\in\{0,1,2...h\}\}$. $w$ and $h$ are the width and height of the image counted in unit lengths. And a set of target keypoints are also defined in the same image plane and denoted as $\{\textbf{k}=(x,y)\}$.

It is worth noting that the size of the sample points defined here is infinitely small and the size of the images' semantically meaningful area is calculated with the unit length. As a result, the image size (i.e., $w$ for width and $h$ for height) we discussed following is different from the resolution of the image matrix, which is widely used for defining the image size in common sense. Formally, the relationship between them is as follow:
\begin{equation}
    \begin{split}
     w = w^p - 1\\
     h = h^p - 1
    \end{split}
\end{equation}
where $w^p$ and $h^p$ are the width and height of the image matrix counted in pixels. We use superscript $p$ to discriminate the variables counted in pixel from those measured in unit length.

\begin{figure}[h]
    \centering
    \includegraphics[width=1.0\hsize]{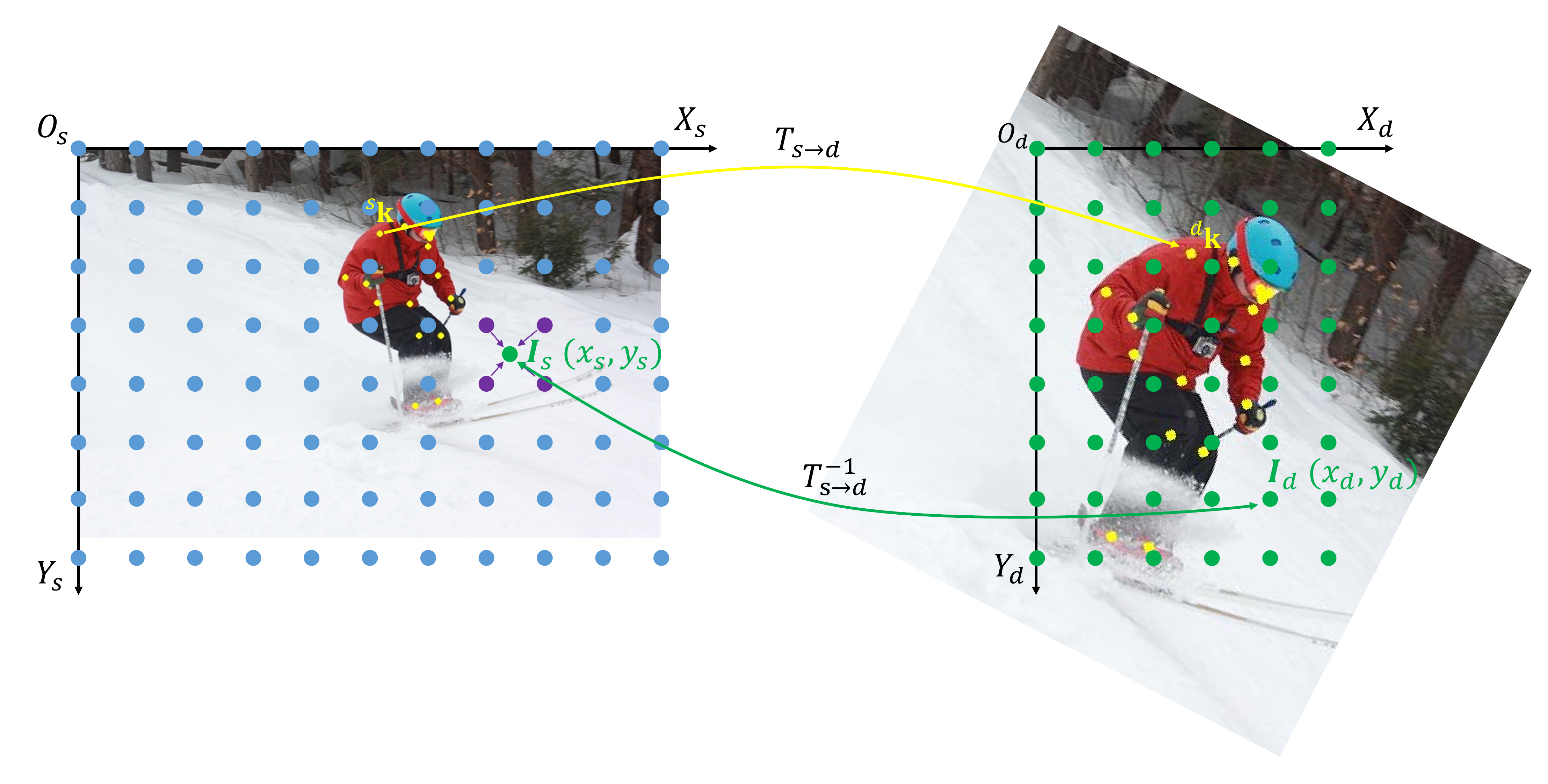}
    \caption{The illustration of the coordinate system transformation in human pose estimation problem. }
    \label{fig:concept_of_coordinate_transformation}
\end{figure}

\subsubsection{The Concept of Coordinate System Transformation.}
The \textit{coordinate system transformation} in human pose estimation can be generally formulated as the data description transformation from the source coordinate system into the destination coordinate system. As illustrated in Figure~\ref{fig:concept_of_coordinate_transformation}, we label the source coordinate system with subscript $s$ as $O_s-X_sY_s$ and the target coordinate system as $O_d-X_dY_d$. Then the transformation of keypoint coordinates can be formulated as:
\begin{equation}
\label{eq:transformation_kps}
    \textbf{k}_d = T_{s\rightarrow d}\textbf{k}_s
\end{equation}
where $T_{s\rightarrow d}$ is the coordinate system transformation matrix from the source coordinate system to the destination coordinate system. And the transformation of the contents in image matrix can be formulated as:
\begin{equation}
    \label{eq:transformation_img}
    \textbf{I}_d(\textbf{p}_d) = \textbf{I}_s(T_{s\rightarrow d}^{-1}\textbf{p}_d)
\end{equation}
where $T_{s\rightarrow d}^{-1}$ is the inverse of $T_{s\rightarrow d}$. Equation~\ref{eq:transformation_img} means that we make the image constant semantically aligned with the annotated keypoints in the destination coordinate system by setting the color of position $\textbf{p}_d$ the same as that in the source image at position $T_{s\rightarrow d}^{-1}\textbf{p}_d$. The results of backtracking $T_{s\rightarrow d}^{-1}\textbf{p}_d$ are usually not integers, and thus, $I_s(T_{s\rightarrow d}^{-1}\textbf{p}_d)$ should be calculated by bilinear interpolation with the valid surrounding points (i.e., the purple points in Figure~\ref{fig:concept_of_coordinate_transformation}). As we only have a sampling result (i.e., the image matrix) of the image plane, interpolation is the optimal way to reduce the precision degradation in image transformation, but can not thoroughly remedy it. Thus, as the precision degradation of interpolation is irreversible and cumulative, we have a principle that the less interpolation done in the data processing pipeline is the better in designing coordinate system transformation pipelines.

\subsubsection{The Targets of Unbias.}
\textit{Unbias} is a target in coordinate system transformation designing, which contains two aspect: One is to keep the semantical alignment after performing transformations. Semantical alignment means that the positional relativeness  between different data (i.e., images and keypoint positions) is unchanged (e.g., the annotated position of nose in destination space is still exactly located upon the nose in the image described in destination space). This is guaranteed by keeping the transformation matrix the same in both Equation~\ref{eq:transformation_kps} and Equation~\ref{eq:transformation_img}.

Another aspect is to make the predicting result exactly aligned with the ground truth under the assumption that the network has a perfect learning ability. In other words, we hope the network's learning ability to be the unique source of precision degradation, and there are no defects in the design of the coordinate transformation pipeline will cause precision degradation. In the following, we will detail our unbiased coordinate system transformation pipeline and prove its unbiased property.

\subsubsection{Coordinate System Transformation in Elementary Operations.}
Coordinate system transformations in human pose estimation derives from some elementary operations like cropping, resizing, rotating and flipping.

\begin{figure}[h]
    \centering
    \includegraphics[width=0.8\hsize]{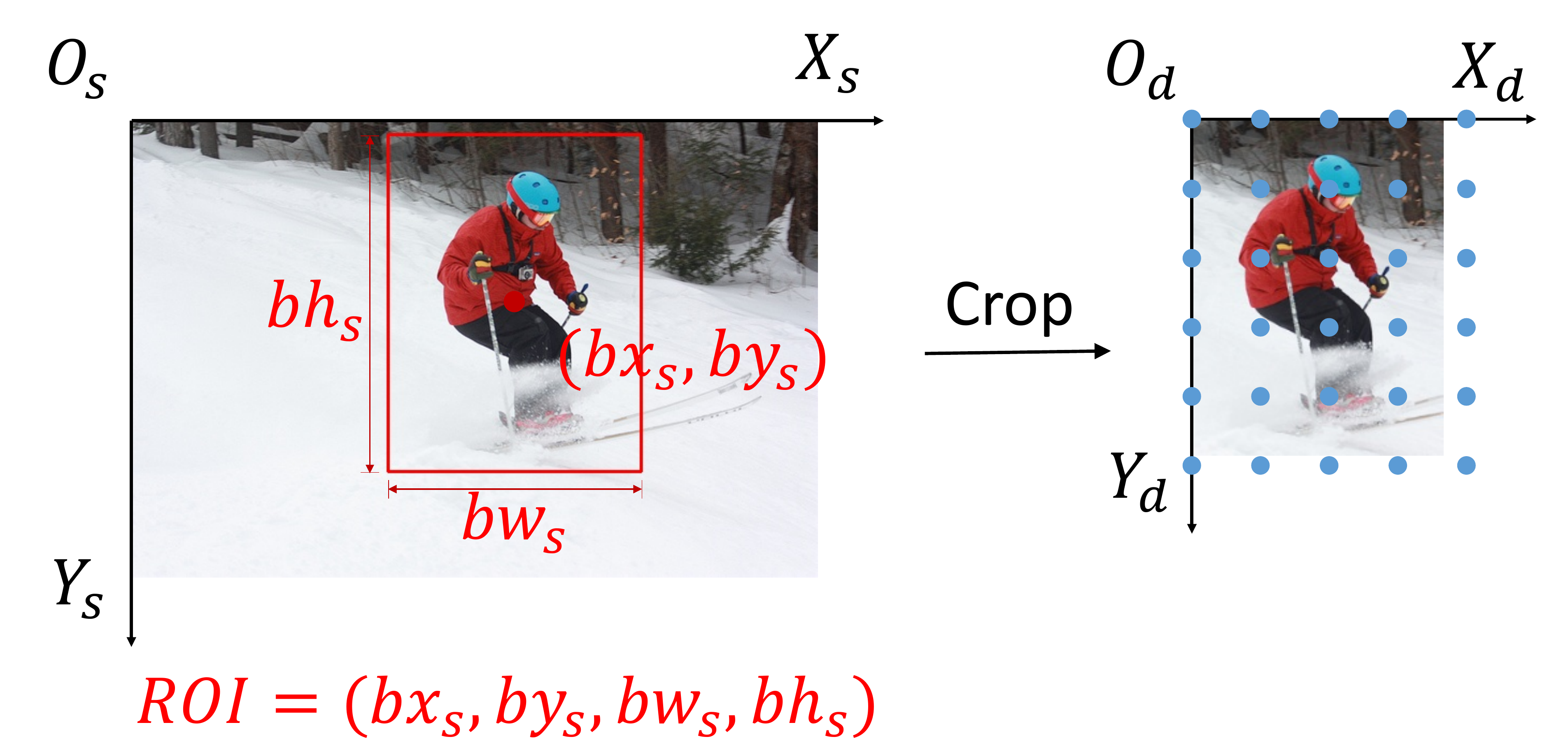}
    \caption{The coordinate system transformation in cropping operation. }
    \label{fig:crop}
\end{figure}

\textbf{Cropping}, as illustrated in Figure~\ref{fig:crop}, is conducted according to a specific Region of Interest (ROI) defined in the source coordinate system $ROI = (bx_s,by_s,bw_s,bh_s)$, where $(bx_s,by_s)$ denotes its center position and $(bw_s,bh_s)$ denotes its width and height. The destination coordinate system can be obtained by moving the origin of the source coordinate system to the upper left corner of the ROI. Thus, the transformation matrix should be designed as:
\begin{equation}
    T_{crop}(ROI) =
    \begin{bmatrix} 1 & 0 & -bx_s+0.5bw_s \\
                    0 & 1 & -by_s+0.5bh_s \\
                    0 & 0 &1 \end{bmatrix}
\end{equation}

\begin{figure}[h]
    \centering
    \includegraphics[width=0.88\hsize]{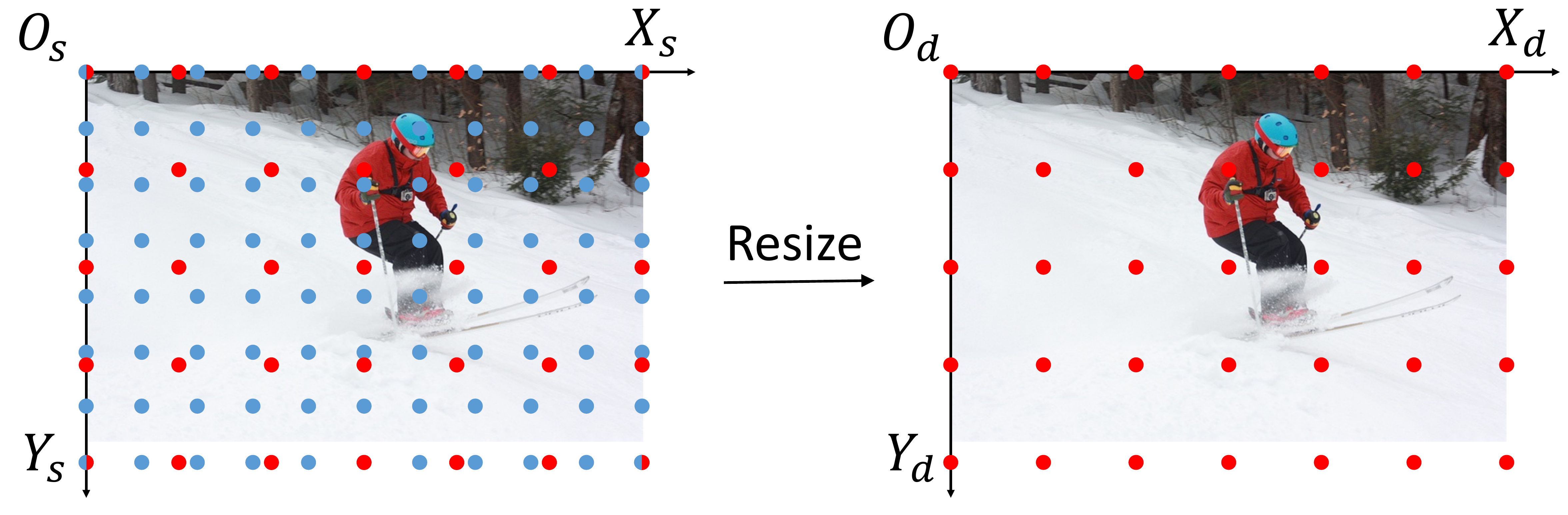}
    \caption{The coordinate system transformation in resizing operation. }
    \label{fig:resize}
\end{figure}

\textbf{Resizing}, as illustrated in Figure~\ref{fig:resize}, changes the sampling strategy only and keep the semantic constant of the image the same as the source. We make the four corner sample point semantically align with the source four corner sample point and let the other sample points evenly distributed among the area dividing by the four corners. Thus the only thing that changes is the unit length of the coordinate system and the transformation matrix should be designed as:
\begin{equation}
    T_{resize}(w_s,h_s,w_d,h_d) =
    \begin{bmatrix} \frac{w_d}{w_s} & 0 & 0 \\
                    0 & \frac{h_d}{h_s} & 0 \\
                    0 & 0 &1 \end{bmatrix}
\end{equation}

\begin{figure}[h]
    \centering
    \includegraphics[width=0.88\hsize]{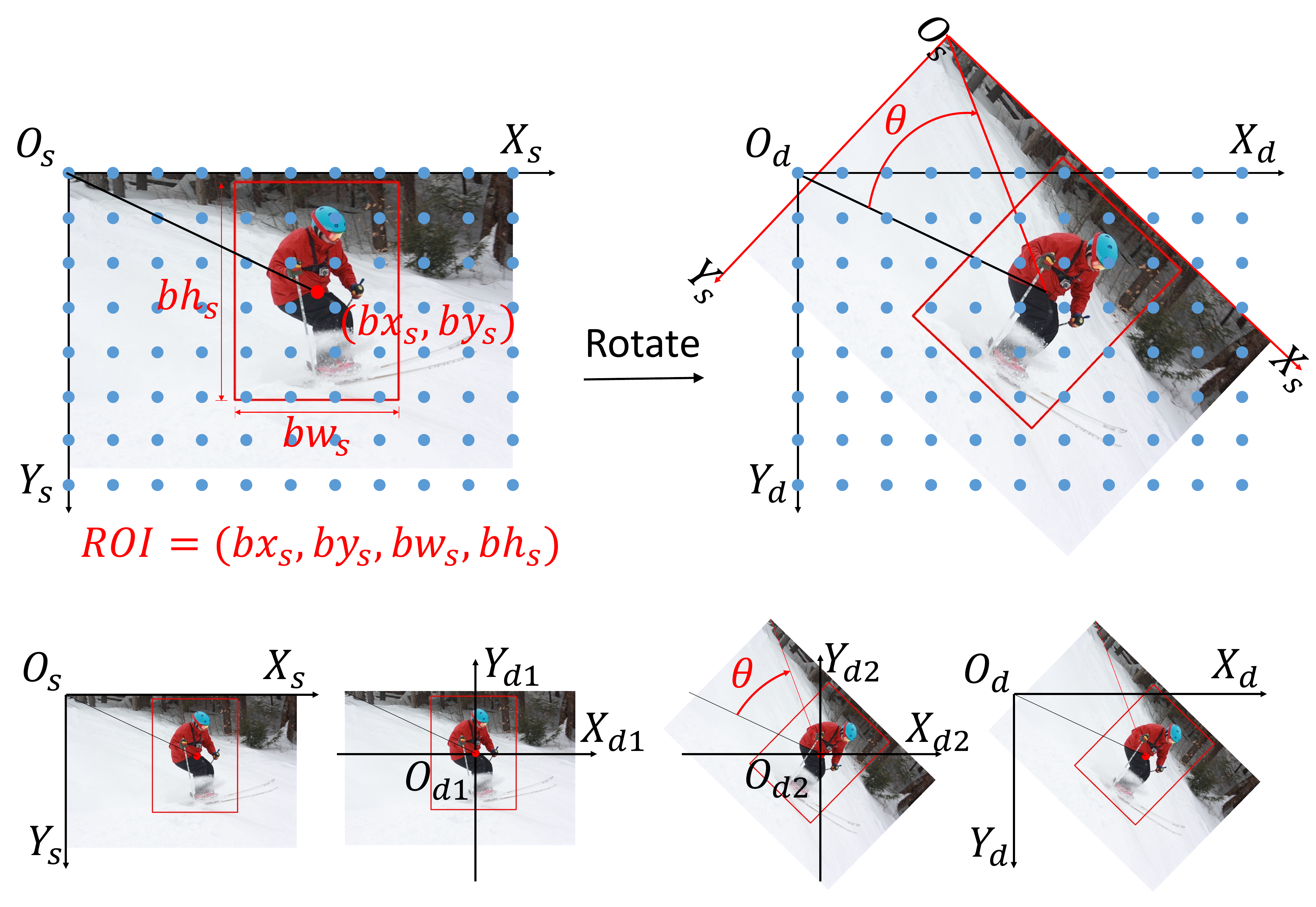}
    \caption{The coordinate system transformation in rotating operation. }
    \label{fig:rotate}
\end{figure}

\textbf{Rotating}, as illustrated in Figure~\ref{fig:rotate}, is conducted according to a rotation center which is always set as the center of a specific ROI instead of the origin of the coordinate system. This design aims at keeping the center position of ROI unchanged (i.e., $(bx_s,by_s)=(bx_d,by_d)$). For example, the ROI refers to the bounding boxes of human instances in top-down paradigm and the whole image in bottom-up paradigm. So, the transformation matrix should be designed as the combination of three elementary transformations:

\begin{equation}
    \label{eqa:Oo2Oi}
    \begin{split}
     &T_{rot}(\theta,ROI) \\
    =&T_{d2\rightarrow d}T_{d1\rightarrow d2}T_{s\rightarrow d1}\\
    =&\begin{bmatrix} 1  & 0 & bx_s \\
                    0 & -1 & by_s\\
                    0& 0 &1 \end{bmatrix}\begin{bmatrix} \cos\theta  & \sin\theta & 0 \\
                    -\sin\theta & \cos\theta & 0\\
                    0& 0 &1 \end{bmatrix}\begin{bmatrix} 1  & 0 & -bx_s \\
                    0 & -1 & by_s\\
                    0& 0 &1 \end{bmatrix}\\
    =&\begin{bmatrix} \cos\theta  & -\sin\theta & -bx_s\cos\theta+by_s\sin\theta+bx_s \\
                      \sin\theta  &  \cos\theta & -bx_s\sin\theta-by_s\cos\theta+by_s \\
                    0& 0 &1 \end{bmatrix}
    \end{split}
\end{equation}

\begin{figure}[h]
    \centering
    \includegraphics[width=0.88\hsize]{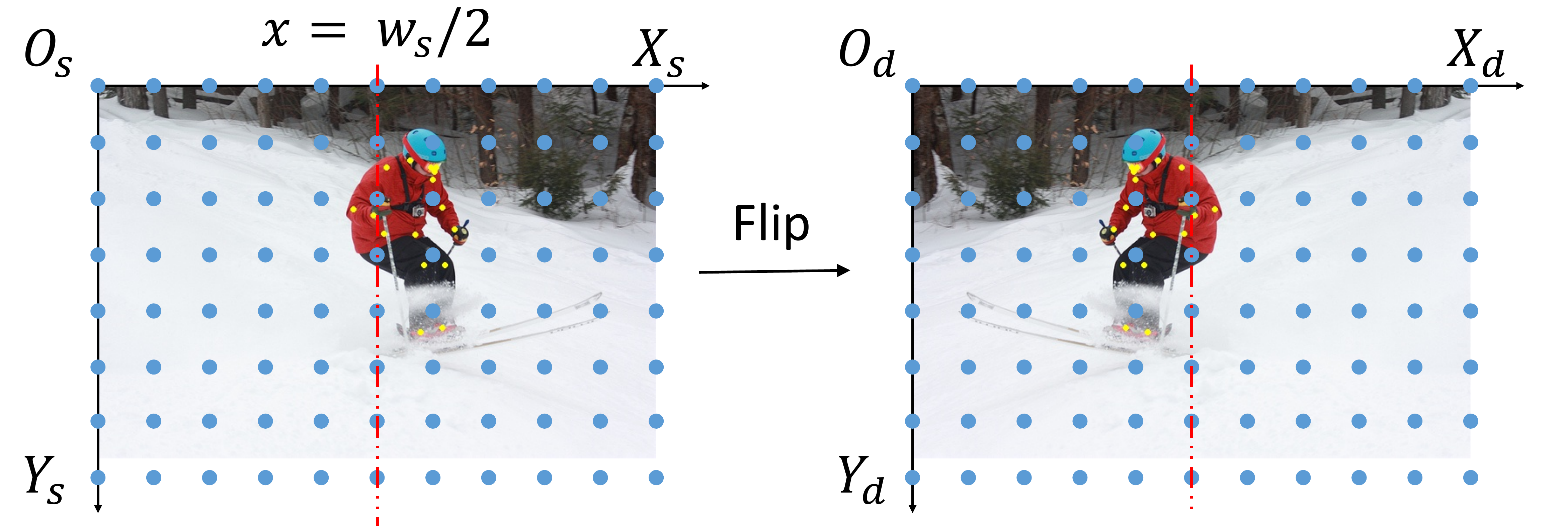}
    \caption{The coordinate system transformation in flipping operation. }
    \label{fig:flip}
\end{figure}
\textbf{Flipping}, as illustrated in Figure~\ref{fig:flip}, generally takes $x = w_s/2$ as the mirror and horizontally exchanges the images' content. So, the transformation matrix should be designed as:
\begin{equation}
    T_{flip}(w_s) =
    \begin{bmatrix} -1 & 0 & w_s \\
                    0 & 1 & 0 \\
                    0 & 0 &1 \end{bmatrix}
\end{equation}

\subsubsection{Common Coordinate System Transformation}
\begin{figure*}[t]
    \centering
    \includegraphics[width=1.0\hsize]{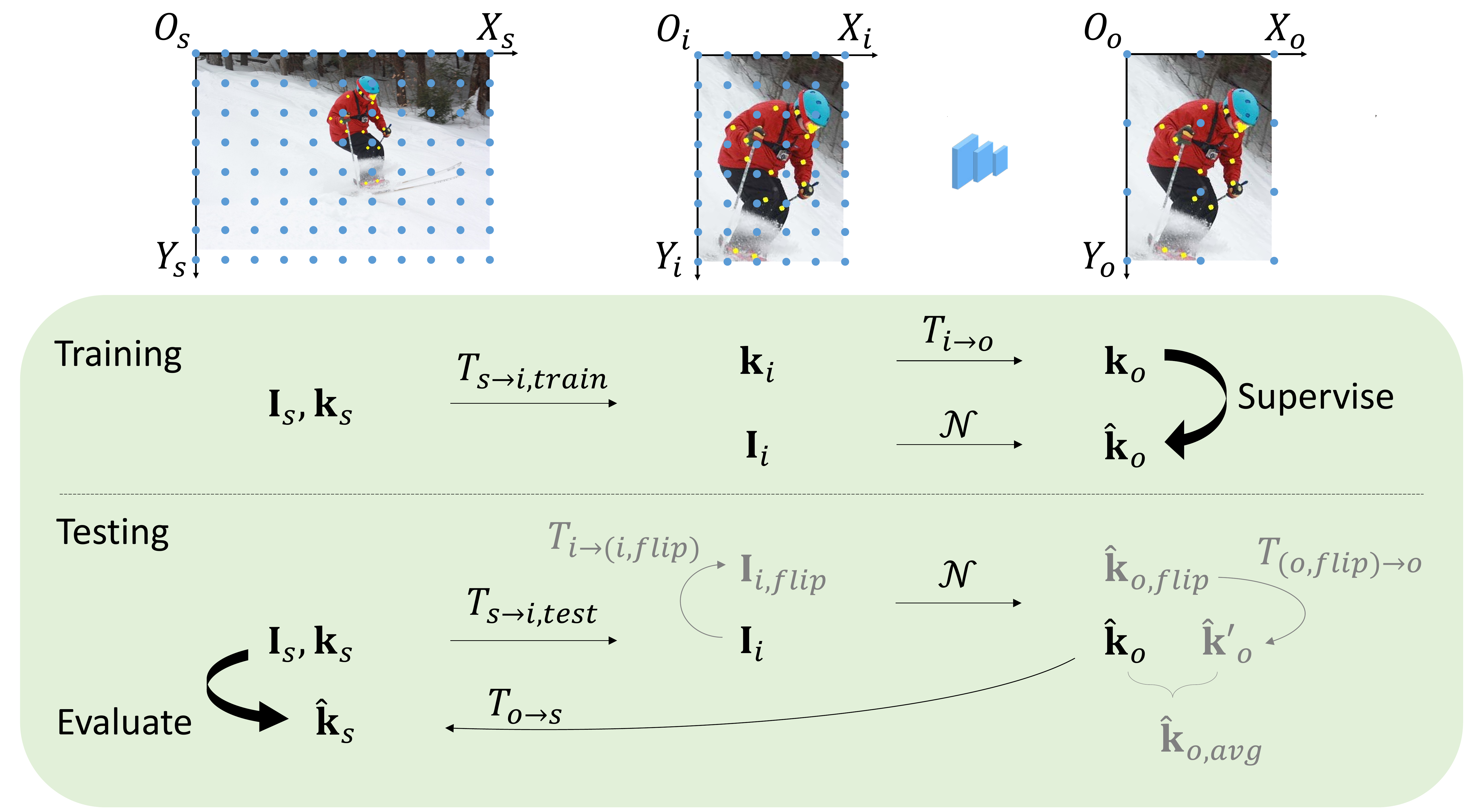}
    \caption{The illustration of the common coordinate system transformation in human pose estimation problem. Three coordinate system are involved: source image coordinate system $O_s\text{-}X_sY_s$, network input coordinate system $O_i\text{-}X_iY_i$ and network output coordinate system $O_o\text{-}X_oY_o$. }
    \label{fig:common_coordinate_transformation_train}
\end{figure*}

In human pose estimation as illustrated in Figure~\ref{fig:common_coordinate_transformation_train}, there are three coordinate systems are involved: source image coordinate systems denoted as $O_s\text{-}X_sY_s$ with subscript $s$ corresponding to the source image with a size of $(w_s,h_s)$, network input coordinate systems denoted as $O_i\text{-}X_iY_i$ with subscript $i$ corresponding to the network input with a size of $(w_i,h_i)$, and network output coordinate systems denoted as $O_o\text{-}X_oY_o$ with subscript $o$ corresponding to the network output with a size of $(w_o,h_o)$.

During the training process, the data is firstly transformed from the source image coordinate systems into the network input coordinate systems according to a specific ROI $(bx_s,by_s,bw_s,bh_s)$ and a rotation angle $\theta$. Some elementary operations are conducted orderly:
\begin{equation}
    \begin{split}
    &T_{flip}(w_i)=\begin{bmatrix} -1 & 0 & w_i \\
                    0 & 1 & 0 \\
                    0 & 0 &1 \end{bmatrix}\\
    &T_{rot}(\theta,(0.5w_i,0.5h_i,w_i,h_i))\\
    &=\begin{bmatrix} \cos\theta  & -\sin\theta & -0.5w_i\cos\theta+0.5h_i\sin\theta+0.5w_i \\
                      \sin\theta  &  \cos\theta & -0.5w_i\sin\theta-0.5h_i\cos\theta+0.5h_i \\
                    0& 0 &1 \end{bmatrix}\\
    &T_{resize}(bw_s,bh_s,w_i,h_i)=\begin{bmatrix} \frac{w_i}{bw_s} & 0 & 0 \\
                    0 & \frac{h_i}{bh_s} & 0 \\
                    0 & 0 &1 \end{bmatrix}\\
    &T_{crop}(bx_s,by_s,bw_s,bh_s)=\begin{bmatrix} 1 & 0 & -bx_s+0.5bw_s \\
                    0 & 1 & -by_s+0.5bh_s \\
                    0 & 0 &1 \end{bmatrix}
    \end{split}
\end{equation}
Then we have the combined transformation:
\begin{equation}
\label{eq:source2input}
    \begin{split}
    &\textbf{k}_i = T_{s\rightarrow i,train}\textbf{k}_s \\
    &\textbf{I}_i(\textbf{p}_i) = I_s(T_{s\rightarrow i,train}^{-1}\textbf{p}_i)\\
    &T_{s\rightarrow i,train} = T_{flip}T_{rot}T_{resize}T_{crop}
    \end{split}
\end{equation}
Equation~\ref{eq:source2input} integrates not only the necessary transformations likes cropping and resizing, but also the optional augmentations (i.e., $T_{flipping}$ for random flipping, $T_{rotating}$ for random rotating, $T_{cropping}$ for half body and random cropping.) used in human pose estimator training. Cropping and Resizing are necessary, while flipping and rotating are optional. The image matrixes in network input space are set as the network input and we have the inference results in the network output space:
\begin{equation}
\label{eq:networkpredict}
    \hat{\textbf{k}}_o = \mathcal{N}(\textbf{I}_i)
\end{equation}
where $\mathcal{N}$ denotes the networks. The annotation is simultaneously transformed from the network input space into the network output space by a simple resizing operation:
\begin{equation}
\label{eq:i2o}
    \begin{split}
    &\textbf{k}_o = T_{i\rightarrow o}\textbf{k}_i \\
    &T_{i\rightarrow o} = T_{resize}\\
    &T_{resize}(w_i,h_i,w_o,h_o)=\begin{bmatrix} \frac{w_o}{w_i} & 0 & 0 \\
                    0 & \frac{h_o}{h_i} & 0 \\
                    0 & 0 &1 \end{bmatrix}\\
    \end{split}
\end{equation}
And $\textbf{k}_o$ in the network output space serves as the supervision:
\begin{equation}
    \label{eq:loss}
    Loss = ||\hat{\textbf{k}}_o-\textbf{k}_o||
\end{equation}
The networks are optimized in the training process and as an ideal result, we have:
\begin{equation}
\label{eq:assumption}
    \begin{split}
    &Loss = ||\hat{\textbf{k}}_o-\textbf{k}_o|| = 0\\
    &\mathcal{N}(\textbf{I}_i) = \hat{\textbf{k}}_o = \textbf{k}_o = T_{i\rightarrow o}\textbf{k}_i
    \end{split}
\end{equation}
which means that the network learns not only the reflection from image matrixes $\textbf{I}_i$ to keypoint positions $\textbf{k}_i$, but also the reflection of the transformation $T_{i\rightarrow o}$ defined in Equation~\ref{eq:i2o}.

In testing process, only the image matrixes are transformed from the source image coordinate systems into the network input coordinate system with the necessary elementary transformations, which should be in line with those in the training process:
\begin{equation}
\label{eq:s2itest}
    \begin{split}
    &\textbf{I}_i(\textbf{p}_i) = \textbf{I}_s(T_{s\rightarrow i,test}^{-1}\textbf{p}_i)\\
    &T_{s\rightarrow i,test} = T_{resize}T_{crop}\\
    &T_{resize}(bw_s,bh_s,w_i,h_i)=\begin{bmatrix} \frac{w_i}{bw_s} & 0 & 0 \\
                    0 & \frac{h_i}{bh_s} & 0 \\
                    0 & 0 &1 \end{bmatrix}\\
    &T_{crop}(bx_s,by_s,bw_s,bh_s)=\begin{bmatrix} 1 & 0 & -bx_s+0.5bw_s \\
                    0 & 1 & -by_s+0.5bh_s \\
                    0 & 0 &1 \end{bmatrix}
    \end{split}
\end{equation}
Then the network outputs in Equation~\ref{eq:networkpredict} are transformed back to the source image coordinate systems by inverse transformations:
\begin{equation}
\label{eq:o2stest}
    \begin{split}
    &\hat{\textbf{k}}_s = T_{o\rightarrow s}\hat{\textbf{k}}_o\\
    &T_{o\rightarrow s} = T_{crop}T_{resize}\\
    &T_{resize}(w_o,h_o,bw_s,bh_s)=\begin{bmatrix} \frac{bw_s}{w_o} & 0 & 0 \\
                    0 & \frac{bh_s}{h_o} & 0 \\
                    0 & 0 &1 \end{bmatrix}\\
    &T_{crop}(0.5w_s-bx_s+0.5bw_s,\\
    &0.5y_s-by_s+0.5bh_s,w_s,h_s)\\
    &=\begin{bmatrix} 1 & 0 & bx_s-0.5bw_s \\
                    0 & 1 & by_s-0.5bh_s \\
                    0 & 0 &1 \end{bmatrix}
    \end{split}
\end{equation}
With Equation~\ref{eq:assumption} as assumption and taking Equation~\ref{eq:i2o}, Equation~\ref{eq:s2itest}, Equation~\ref{eq:o2stest} into consideration, we have the following identical relation:
\begin{equation}
    \label{eq:inference1}
    \begin{split}
    \hat{\textbf{k}}_s  &= T_{o\rightarrow s}\hat{\textbf{k}}_o\\
                        &= T_{o\rightarrow s}T_{i\rightarrow o}\textbf{k}_i\\
                        &= T_{o\rightarrow s}T_{i\rightarrow o}T_{s\rightarrow i, test}\textbf{k}_s\\
                        &= \begin{bmatrix} 1 & 0 & bx_s-0.5bw_s \\
                    0 & 1 & by_s-0.5bh_s \\
                    0 & 0 &1 \end{bmatrix}\begin{bmatrix} \frac{bw_s}{w_o} & 0 & 0 \\
                    0 & \frac{bh_s}{h_o} & 0 \\
                    0 & 0 &1 \end{bmatrix}\begin{bmatrix} \frac{w_o}{w_i} & 0 & 0 \\
                    0 & \frac{h_o}{h_i} & 0 \\
                    0 & 0 &1 \end{bmatrix}\\
                    &\begin{bmatrix} \frac{w_i}{bw_s} & 0 & 0 \\
                    0 & \frac{h_i}{bh_s} & 0 \\
                    0 & 0 &1 \end{bmatrix}\begin{bmatrix} 1 & 0 & -bx_s+0.5bw_s \\
                    0 & 1 & -by_s+0.5bh_s \\
                    0 & 0 &1 \end{bmatrix}\textbf{k}_s\\
                        &= \textbf{k}_s
    \end{split}
\end{equation}
This inference prove that the result in the source image space is exactly equal to the ground truth, which means that the data transformation pipeline designed above is unbiased and no systematic error would be involved.

When flipping ensemble is used in testing process, the flipped image is obtained by performing flipping transformation in the network input space:
\begin{equation}
\label{eq:flipinput}
    \begin{split}
    &\textbf{I}_{i,flip}(\textbf{p}_{i,flip}) = \textbf{I}_{i}(T_{i\rightarrow (i,flip)}^{-1}\textbf{p}_{i,flip})\\
    &T_{i\rightarrow (i,flip)} = T_{flip}(w_i) =    \begin{bmatrix} -1 & 0 & w_i \\
                                                                        0 & 1 & 0 \\
                                                                        0 & 0 &1 \end{bmatrix}
    \end{split}
\end{equation}
Then we have the network prediction $\hat{\textbf{k}}_{o,flip} = \mathcal{N}(\textbf{I}_{i,flip})$ which is subsequently flipped back in the network output space:
\begin{equation}
\label{eq:flipoutput}
    \begin{split}
    &\hat{\textbf{k}}_{o}' = T_{(o,flip)\rightarrow o}\hat{\textbf{k}}_{o,flip}\\
    &T_{(o,flip)\rightarrow o} = T_{flip}(w_o) =    \begin{bmatrix}   -1 & 0 & w_o \\
                                                                            0 & 1 & 0 \\
                                                                            0 & 0 &1 \end{bmatrix}
    \end{split}
\end{equation}
with Equation~\ref{eq:assumption} as assumption and taking Equation~\ref{eq:i2o}, Equation~\ref{eq:flipinput} and Equation~\ref{eq:flipoutput} into consideration, we have the following identical relation:
\begin{equation}
\label{eq:inference2}
    \begin{split}
    \hat{\textbf{k}}_o' &= T_{(o,flip)\rightarrow o}\hat{\textbf{k}}_{o,flip}\\
                        &= T_{(o,flip)\rightarrow o}T_{(i,flip)\rightarrow (o,flip)}\textbf{k}_{i,flip} \\
                        &= T_{(o,flip)\rightarrow o}T_{(i,flip)\rightarrow (o,flip)}T_{i \rightarrow (i,flip)}\textbf{k}_{i} \\
                        &= T_{flip}(w_o)T_{resize}(w_i,h_i,w_o,h_o)T_{flip}(w_i)\textbf{k}_{i} \\
                        &= \begin{bmatrix} -1 & 0 & w_o \\
                    0 & 1 & 0 \\
                    0 & 0 &1 \end{bmatrix}\begin{bmatrix} \frac{w_o}{w_i} & 0 & 0 \\
                    0 & \frac{h_o}{h_i} & 0 \\
                    0 & 0 &1 \end{bmatrix}\begin{bmatrix} -1 & 0 & w_i \\
                    0 & 1 & 0 \\
                    0 & 0 &1 \end{bmatrix}\textbf{k}_{i}\\
                        &= \begin{bmatrix} \frac{w_o}{w_i} & 0 & 0 \\
                    0 & \frac{h_o}{h_i} & 0 \\
                    0 & 0 &1 \end{bmatrix}\textbf{k}_i\\
                        &= T_{i\rightarrow o}\textbf{k}_i\\
                        &= \hat{\textbf{k}}_o
    \end{split}
\end{equation}
This inference prove that, in the network output space, the results from flipped images are aligned with those from the origin images. By taking Equation~\ref{eq:inference1} into consideration, the results from flipped images in the source image space are also aligned with the ground truths and no systematic error would be involved. The establish of Equation~\ref{eq:inference1} and Equation~\ref{eq:inference2} guarantees the unbiased property in the coordinate system transformation pipeline. They will be used as the guideline for checking biased coordinate system transformation pipelines in the following subsection.
\subsubsection{Diagnosis of the Biased Coordinate System Transformation}
\label{sec:decst}
In most state-of-the-arts \cite{CPN,SBNet,MSPN,HRNet,Higher}, the bias problem in coordinate system transformation pipeline derives from using resolution $(w_s^p,h_s^p)$ counted in pixels instead of size $(w_s,h_s)$ measured in unit length when performing resizing transformation. As a consequence, it changes Equation~\ref{eq:inference1} and Equation~\ref{eq:inference2} into:
\begin{equation}
    \label{eq:inference3}
    \begin{split}
    \hat{\textbf{k}}_s  &= T_{o\rightarrow s}\hat{\textbf{k}}_o\\
                        &= T_{o\rightarrow s}T_{i\rightarrow o}\textbf{k}_i\\
                        &= T_{o\rightarrow s}T_{i\rightarrow o}T_{s\rightarrow i, test}\textbf{k}_s\\
                        &= \begin{bmatrix} 1 & 0 & bx_s-0.5bw_s \\
                    0 & 1 & by_s-0.5bh_s \\
                    0 & 0 &1 \end{bmatrix}\begin{bmatrix} \frac{bw_s}{w_o^p} & 0 & 0 \\
                    0 & \frac{bh_s}{h_o^p} & 0 \\
                    0 & 0 &1 \end{bmatrix}\begin{bmatrix} \frac{w_o^p}{w_i^p} & 0 & 0 \\
                    0 & \frac{h_o^p}{h_i^p} & 0 \\
                    0 & 0 &1 \end{bmatrix}\\
                    &\begin{bmatrix} \frac{w_i^p}{bw_s} & 0 & 0 \\
                    0 & \frac{h_i^p}{bh_s} & 0 \\
                    0 & 0 &1 \end{bmatrix}\begin{bmatrix} 1 & 0 & -bx_s+0.5bw_s \\
                    0 & 1 & -by_s+0.5bh_s \\
                    0 & 0 &1 \end{bmatrix}\textbf{k}_s\\
                        &= \textbf{k}_s
    \end{split}
\end{equation}
\begin{equation}
\label{eq:inference4}
    \begin{split}
    \hat{\textbf{k}}_o' &= T_{(o,flip)\rightarrow o}\hat{\textbf{k}}_{o,flip}\\
                        &= T_{(o,flip)\rightarrow o}T_{(i,flip)\rightarrow (o,flip)}\textbf{k}_{i,flip} \\
                        &= T_{(o,flip)\rightarrow o}T_{(i,flip)\rightarrow (o,flip)}T_{i \rightarrow (i,flip)}\textbf{k}_{i} \\
                        &= T_{flip}(w_o)T_{resize}(w_i^p,h_i^p,w_o^p,h_o^p)T_{flip}(w_i)\textbf{k}_{i} \\
                        &= \begin{bmatrix} -1 & 0 & w_o \\
                    0 & 1 & 0 \\
                    0 & 0 &1 \end{bmatrix}\begin{bmatrix} \frac{w_o^p}{w_i^p} & 0 & 0 \\
                    0 & \frac{h_o^p}{h_i^p} & 0 \\
                    0 & 0 &1 \end{bmatrix}\begin{bmatrix} -1 & 0 & w_i \\
                    0 & 1 & 0 \\
                    0 & 0 &1 \end{bmatrix}\textbf{k}_{i}\\
                        &= \begin{bmatrix} \frac{w_o^p}{w_i^p} & 0 & \frac{w_o^p}{w_i^p}-1 \\
                    0 & \frac{h_o^p}{h_i^p} & 0 \\
                    0 & 0 &1 \end{bmatrix}\textbf{k}_i\\
                        &= \begin{bmatrix} 1 & 0 & \frac{w_o^p}{w_i^p}-1 \\
                    0 & 1 & 0 \\
                    0 & 0 &1 \end{bmatrix}\begin{bmatrix} \frac{w_o^p}{w_i^p} & 0 & 0 \\
                    0 & \frac{h_o^p}{h_i^p} & 0 \\
                    0 & 0 &1 \end{bmatrix}\textbf{k}_i\\
                        &= \begin{bmatrix} 1 & 0 & \frac{1-s}{s} \\
                    0 & 1 & 0 \\
                    0 & 0 &1 \end{bmatrix}\hat{\textbf{k}}_o\\
    \end{split}
\end{equation}
where $s = w_i^p/w_o^p$ is the stride factor for describing the size variation of network features. Here, $\hat{\textbf{k}}_s$ is still equal to $\textbf{k}_s$, indicating that the aforementioned modification will not change the unbiased property in coordinate system transformation pipeline $T_{o\rightarrow s}T_{i\rightarrow o}T_{s\rightarrow i, test}$ and should have no effect on the precision of the predicted results. However when flipping ensemble is adopted in testing process, $\hat{\textbf{k}}_{o}'$ is not exactly aligned with $\hat{\textbf{k}}_o$, and there is an offset of $\frac{1-s}{s}$ in $O_o\text{-}X_o$ direction. Taking $\hat{\textbf{k}}_o$ as reference, $\frac{1-s}{s}$ is the predicting error of result $\hat{\textbf{k}}_{o}'$ in network output space. If we directly average $\hat{\textbf{k}}_{o}'$ and $\hat{\textbf{k}}_o$ as done in most existing works:
\begin{equation}
    \hat{\textbf{k}}_{o,avg} = \frac{\hat{\textbf{k}}_{o}' +\hat{\textbf{k}_o}}{2} = \begin{bmatrix} 1 & 0 & \frac{1-s}{2s} \\
                                                                                        0 & 1 & 0 \\
                                                                                        0 & 0 &1 \end{bmatrix}\hat{\textbf{k}}_o\\
\end{equation}
the final error in $O_o\text{-}X_o$ direction is:
\begin{equation}
    \label{eq:errornoshift}
    e(x)_o=|x(\hat{\textbf{k}}_{o,avg}) - x(\hat{\textbf{k}}_o)| = |\frac{1-s}{2s}|=0.375|_{s=4}
\end{equation}
where $\hat{\textbf{k}}_o$ is regarded as ground truth as it has been proved unbiased by Equation~\ref{eq:inference3}. The magnitude of this predicting error is so large that the performance will be degraded by a considerable margin. In state-of-the-arts, there are some empirical remedies for this error, which can be classified into two categories: direct compensation or using higher resolution.

As the error $|\frac{1-s}{2s}|$ has a fixed scale which is determined by the stride factor, direct compensation is effective, being the remedy in most state-of-the-arts top-down methods \cite{CPN, MSPN, SBNet, HRNet, DARK}. For example, SimpleBaseline \cite{SBNet}, HRNet \cite{HRNet} and DarkPose \cite{DARK} empirically shift the result from flipped image by one pixel in $O_o\text{-}X_o$ direction before performing the averaging operation to suppress this error:
\begin{equation}
    \hat{\textbf{k}}_{o,avg} = \frac{\begin{bmatrix} 1 & 0 & 1 \\
                                            0 & 1 & 0 \\
                                            0 & 0 &1 \end{bmatrix}\hat{\textbf{k}}_{o}' + \hat{\textbf{k}}_o}{2} = \begin{bmatrix} 1 & 0 & \frac{1}{2s} \\
                                            0 & 1 & 0 \\
                                            0 & 0 &1 \end{bmatrix}\hat{\textbf{k}}_o\\
\end{equation}
In this way, the final error can be reduced to
\begin{equation}
    \label{eq:errorshift1pixel}
    e(x)_o' = |\frac{1}{2s}|=0.125|_{s=4}
\end{equation}
where $e(x)_o'< e(x)_o$ when $s>2$, which makes sense in most existing top-down methods with a stride factor of $4$ \cite{CPN, MSPN, SBNet, HRNet, DARK}. Intuitively, as a result of reasoning, an extra compensation for $e(x)_o'$ in network output space can make the result of existing work more accurate. We will verify this in ablation study.

Besides, when mapping $e(x)_o'$ back to source image coordinate system ($O_s\text{-}X_sY_s$) with Equation~\ref{eq:o2stest}, we have:
\begin{equation}
\label{eq:errorinsourcespace}
    e(x)_s' = |\frac{1}{2s}\times\frac{bw_s}{w_o^p}| = |\frac{bw_s}{2w_i^p}|
\end{equation}
where $bw_s$ is fixed in inference process. Equation~\ref{eq:errorinsourcespace} means that higher network input resolution can help suppress the predicted error caused by $e(x)_s'$. In other words, the existing top-down methods benefit more from higher input resolution and suffer more accuracy loss from lower input resolution.

Without shifting one pixel in network output space, we have:
\begin{equation}
    e(x)_s = |\frac{s-1}{2s}\times\frac{bw_s}{w_o^p}| = |\frac{bw_s(s-1)}{2w_i^p}|
\end{equation}
which means that both higher input resolution and higher output resolution can help suppress this error. And this contributes the most performance boosting in HigherHRnet \cite{Higher} who empirically proposes to use higher output resolution to pursue high precision at the cost of tremendous latency in both network inference and post processing. By contrast, unbiased data processing provides a free access to achieve similar performance improvement with a low output resolution. Besides of using higher output resolution, HigherHRNet uses another unreported operation that resizes the network output into a a resolution as high as the network input. This operation also coincidentally remedies the error caused by biased coordinate system transformation pipeline and benefit the performance of HigherHRNet structure at the cost of extra latency in post processing. Through ablation study, we will show this operation are gilding the lily when the coordinate system transformation pipeline is unbiased as it will involve extra error and change the distribution of the network output by performing an extra interpolation in resizing operation.

\subsection{Unbiased Keypoint Format Transformation}
\subsubsection{The Concept of Unbiased Keypoint Format Transformation.}
As the coordinate of keypoint is not the superior format for convolutional network study, the intuitively more proper format of heatmap is proposed and quickly has been proved effective. The \textit{keypoint fromat transformation} refers to the transformations between keypoint coordinates and heatmaps which is widely used in state-of-the-art methods. In common sense, \textit{encoding} denotes the transformation from coordinate format into heatmap format, while \textit{decoding} denotes the inverse transformation.
\begin{equation}
    \begin{split}
    &\mathcal{H} = Encoding(\textbf{k})\\
    &\textbf{k} = Decoding(\mathcal{H})
    \end{split}
\end{equation}

Target of \textit{Unbiased} in keypoint format transformation designing is to avoid precision degeneration in the encoding and decoding transformation. As a formulated target, we should have:
\begin{equation}
\label{eq:target}
    \textbf{k} = Decoding(Encoding(\textbf{k}))
\end{equation}

\subsubsection{Unbiased Keypoint Format Transformation.}
In this subsection, we will introduce two unbiased keypoint format transformation paradigm and simultaneously showcase their unbias property.

\textbf{Combined classification and regression format} is inspired by the works in object detection \cite{ren2016faster} where anchors are used to predict bounding boxes, and first proposed in \cite{G-RMI}. We give details introduction here with some modifications. In training process, each annotated keypoint $\textbf{k}=(m,n)$ is encoded through:
\begin{equation}
\label{eq:combinedencoding}
    \begin{split}
    &\mathcal{C}(x,y,m,n)= \begin{cases} 1  & \textbf{if} \ (x-m)^2+(y-n)^2<r^2 \\
                                                    0  & \textbf{otherwise}
    \end{cases}\\
    &\mathcal{X}(x,y,m,n)= m - x\\
    &\mathcal{Y}(x,y,m,n)= n - y\\
    \end{split}
\end{equation}
where $\mathcal{C}$ is the classification heatmap act as the anchor in object detection for preliminarily locate the keypoint. $r$ is a hyper-parameter referring the radius of the area classified as positive. Consist of offset vectors, $\mathcal{X}$ and $\mathcal{Y}$ are the regression heatmap for preserving the residual locating information. Then the loss is designed as:
\begin{equation}
    \begin{split}
    &Loss =  Loss_{cls} + Loss_{reg}\\
    &Loss_{cls} = ||\mathcal{C}-\hat{\mathcal{C}}||\\
    &Loss_{reg} = \mathcal{C} * ||\mathcal{X}-\hat{\mathcal{X}}|| + \mathcal{C} * ||\mathcal{Y}-\hat{\mathcal{Y}}||\\
    &\hat{\mathcal{C}}, \hat{\mathcal{X}}, \hat{\mathcal{Y}} = \mathcal{N}(I)
    \end{split}
\end{equation}
where $\mathcal{C}$ in $Loss_{reg}$ defines the region of interesting, which means that we only need to learn the offset among the area where the classification label is true. The network is optimized in the training process and as a ideal result, we have:
\begin{equation}
\label{eq:assumption1}
    \hat{\mathcal{C}}, \hat{\mathcal{X}}, \hat{\mathcal{Y}} = \mathcal{C}, \mathcal{X}, \mathcal{Y}
\end{equation}

Then in testing processing, the prediction is decoding by:
\begin{equation}
    \begin{split}
    &\hat{\textbf{k}} = \hat{\textbf{k}}_{h} + (\hat{\mathcal{X}}(\hat{\textbf{k}}_{h}),\hat{\mathcal{Y}}(\hat{\textbf{k}}_{h}))\\
    &\hat{\textbf{k}}_{h} = argmax(\hat{\mathcal{C}})
    \end{split}
\end{equation}
where the position of highest response $\hat{\textbf{k}}_{h}$ is located first and is subsequently updated by utilizing the predicted offsets. By taking Equation~\ref{eq:combinedencoding} and Equation~\ref{eq:assumption1} into consideration, we have:
\begin{equation}
    \begin{split}
    \hat{\textbf{k}} &= \hat{\textbf{k}}_{h} + (\hat{\mathcal{X}}(\hat{\textbf{k}}_{h}),\hat{\mathcal{Y}}(\hat{\textbf{k}}_{h}))\\
                     &= (x(\hat{\textbf{k}}_{h}),y(\hat{\textbf{k}}_{h})) + (m-x(\hat{\textbf{k}}_{h}),n-y(\hat{\textbf{k}}_{h}))\\
                     &= (m,n)\\
                     &= \textbf{k}
    \end{split}
\end{equation}
which means that no systematic error is involved in the keypoint format transformation pipeline and the unbiased target in Equation~\ref{eq:target} is achieved.

\textbf{Classification format} is widely used in most state-of-the-arts, where classification heatmap is used only with a gaussian-like distribution:
\begin{equation}
\label{eq:encodinggaussian}
    \mathcal{C}(x,y,m,n)= exp(-\frac{(x-m)^2+(y-n)^2}{2\delta ^2})
\end{equation}
The loss is designed as:
\begin{equation}
    \begin{split}
    &Loss =  ||\mathcal{C}-\hat{\mathcal{C}}||\\
    &\hat{\mathcal{C}}= \mathcal{N}(I)
    \end{split}
\end{equation}
The network is optimized in the training process and as a ideal result, we have:
\begin{equation}
\label{eq:assumption2}
    \hat{\mathcal{C}} = \mathcal{C}
\end{equation}

In testing process, we introduce the decoding method DARK \cite{DARK} who decoding the classification heatmap into keypoint coordinates by searching the center of the gaussian distribution where the first derivative is equal to zero:
\begin{equation}
\label{eq:unbiaseddecodinggaussian}
    \begin{split}
    &\hat{\textbf{k}} = \hat{\textbf{k}}_h - \hat{\mathcal{C}}''(\hat{\textbf{k}}_h)^{-1}\hat{\mathcal{C}}'(\hat{\textbf{k}}_h)\\
    &\hat{\textbf{k}}_{h} = argmax(\hat{\mathcal{C}})
    \end{split}
\end{equation}
where $\hat{\mathcal{C}}'$ and $\hat{\mathcal{C}}''$ are the first order derivative and second order derivative (i.e., Hessian) of $\hat{\mathcal{C}}$. According to \cite{DARK}, the precision degradation caused by Taylor series approximation is negligible and $\hat{\textbf{k}}$ is theoretically close to $\textbf{k}$, which matches the purpose of unbias.

\subsubsection{Analysis of Biased Keypoint Format Transformation.}
We take the keypoint format transformation method used in SimpleBaseline \cite{SBNet}, HRNet \cite{HRNet} and HigherHRNet \cite{Higher} as the example for studying the effect of biased data form transformation. keypoints are encoded into classification heatmap with gaussian distribution as Equation~\ref{eq:encodinggaussian}, but are decoded by the a suboptimal method:
\begin{equation}
    \label{eq:biasdecoding}
    \begin{split}
    &\hat{\textbf{k}} = \hat{\textbf{k}}_h + 0.25 * sign(\hat{\mathcal{C}}'(\hat{\textbf{k}}_{h}))\\
    &\hat{\textbf{k}}_{h} = argmax(\hat{\mathcal{C}})\\
    &sign(x) = \begin{cases}
		      1  & \textbf{if}\  x>0\\
		      -1 & \textbf{otherwise}
	\end{cases}
    \end{split}
\end{equation}
According to the encoding in Equation~\ref{eq:encodinggaussian}, we have
\begin{equation}
    \begin{split}
	&argmax(\hat{\mathcal{C}}) =
    \begin{cases}
		Floor(m)  \!\! & \textbf{if}\ m-Floor(m)<0.5\\
        Ceil(m)  \!\! & \textbf{otherwise}
	\end{cases}\\
    &sign(\hat{\mathcal{C}}'(\hat{\textbf{k}}_{h})) =
    \begin{cases}
		      1  & \textbf{if}\ m-Floor(m)<0.5\\
		      -1 & \textbf{otherwise}
	\end{cases}
    \end{split}
\end{equation}
As an example, predicting coordinate in $O-X$ direction has the distribution of:
\begin{equation}
	\hat{m} =
    \begin{cases}
		Floor(m)+0.25  \!\! & \textbf{if}  \ m-Floor(m)<0.5\\
        Ceil(m) - 0.25  \!\! & \textbf{otherwise}
	\end{cases}
\end{equation}
With the assumption that $\textbf{k}$ is uniformly distributed in the image plane (i.e., both $m-Floor(m)$ and $n-Floor(n)$ are uniformly distributed in interval $[0,1)$), the expected error in each direction is $E(|m-\hat{m}|) = E(|n-\hat{n}|) =1/8 = 0.125$ unit length with a variance of $V(|m-\hat{m}|)= V(|n-\hat{n}|) =1/192\approx0.0052$.

When mapping $E(|m-\hat{m}|)$ back to the source image coordinate system ($O_s\text{-}X_sY_s$) with Equation~\ref{eq:o2stest}, we have:
\begin{equation}
		E(|m_s-\hat{m}_s|) =  E(|m_o-\hat{m}_o|)\times\frac{bw_s}{w_o}
\end{equation}
Considering error $E(|m-\hat{m}|) $ and $E(|n-\hat{n}|) $, the methods with biased data form transformation benefit from higher network output resolution. And this also contributes part of the performance boosting in HigherHRnet \cite{Higher}.

\begin{table*}
\caption{Performance of proposed UDP on COCO \textit{val} set. IPS used in bottom-up paradigm denotes the inference speed of Image Per Second. PPS used in top-down paradigm denotes the inference speed of Person Per Second. $\dagger$ means unreported results in the original paper and trained with official implementation by us.}
\label{tab:val}
\begin{center}
\begin{tabular}{l|l|c|l|lccccc}

\hline
Method                   & Backbone         &Input size    &IPS/PPS                     &AP                     & $\text{AP}^{50}$ & $\text{AP}^{75}$ & $\text{AP}^{\text{M}}$ &$\text{AP}^{\text{L}}$ &AR  \\
\hline
\multicolumn{10}{c}{Bottom-up methods}\\
\hline
HigherHRNet \cite{Higher}& HRNet-W32     &$512\times512$    &0.8                        &64.4                   & -             & -             & 57.1          &75.6           &-             \\
\textbf{+UDP}            & HRNet-W32     &$512\times512$    &\textbf{4.9 ($\times$6.1)} &\textbf{67.0 (+2.6)}   & \textbf{86.2} & \textbf{72.0} & \textbf{60.7} &\textbf{76.7}  &\textbf{71.6}          \\
HigherHRNet \cite{Higher}& HigherHRNet-W32&$512\times512$   &1.1                        &67.1                   & 86.2          & 73.0          & 61.5          &76.1           & 718            \\
\textbf{+UDP}            & HigherHRNet-W32&$512\times512$   &\textbf{2.9 ($\times$2.6)} &\textbf{67.8 (+0.7)}   & \textbf{86.2} & \textbf{72.9} & \textbf{62.2} &\textbf{76.4}  &\textbf{72.4}  \\
HigherHRNet \cite{Higher}$\dagger$&HRNet-W48&$640\times640$ &0.6                        &67.9                  & 86.7              & 74.4         & 62.5          &76.2           &73.0\\
\textbf{+UDP}            &HRNet-W48      &$640\times640$    &\textbf{4.1 ($\times$6.8)} &\textbf{68.9 (+1.0)}         & \textbf{87.3}     & \textbf{74.9}& \textbf{64.1} &\textbf{76.1}  &\textbf{73.5}\\
HigherHRNet \cite{Higher}& HigherHRNet-W48&$640\times640$   &0.75                       &69.9                   & 87.2          & 76.1          & 65.4          &76.4           & -     \\
\textbf{+UDP}            & HigherHRNet-W48&$640\times640$   &\textbf{2.7 ($\times$3.6)} &\textbf{69.9}          & \textbf{87.3} & \textbf{76.2} & \textbf{65.9} &\textbf{76.2}  &\textbf{74.4} \\
\hline
\multicolumn{10}{c}{Bottom-up methods with multi-scale ([$\times$2,$\times$1,$\times$0.5]) test as in HigherHRNet \cite{Higher}}\\
\hline
\textbf{UDP}             & HRNet-W32      &$512\times512$   &-&\textbf{70.4}        & \textbf{88.2} & \textbf{75.8} & \textbf{65.3} &\textbf{77.6}  &\textbf{74.7} \\
HigherHRNet \cite{Higher}& HigherHRNet-W32&$512\times512$   &-&69.9                 & 87.1          & 76.0          & 65.3          &77.0           & -            \\
\textbf{+UDP}            & HigherHRNet-W32&$512\times512$   &-&\textbf{70.2 (+0.3)} & \textbf{88.1} & \textbf{76.2} & \textbf{65.4} &\textbf{77.4}  &\textbf{74.5} \\
HigherHRNet \cite{Higher}$\dagger$&HRNet-W48&$640\times640$ &-&71.6                 & 88.6              & 77.9           & 67.5          &77.8           &76.3\\
\textbf{+UDP}            &HRNet-W48       &$640\times640$   &-&\textbf{71.3 (-0.3)}        & \textbf{89.0}     & \textbf{77.1}  & \textbf{66.9} &\textbf{77.7}  &\textbf{75.7}\\
HigherHRNet \cite{Higher}& HigherHRNet-W48&$640\times640$   &-&72.1                 & 88.4          & 78.2          & 67.8          &78.3           & -            \\
\textbf{+UDP}            & HigherHRNet-W48&$640\times640$   &-&\textbf{71.5 (-0.6)} & \textbf{88.3} & \textbf{77.3} & \textbf{67.9} &\textbf{77.2}  &\textbf{75.9} \\
\hline
\multicolumn{10}{c}{Top-down methods}\\
\hline
Hourglass\cite{Hourglass}  & Hourglass        &$256\times192$    &-&66.9  & -            & -                   & -                  &-                &-   \\
CPN\cite{CPN}              & ResNet-50        &$256\times192$    &-&69.4   & -            & -                   & -                  &-                &-   \\
CPN\cite{CPN}              & ResNet-50        &$384\times288$    &-&71.6   & -                & -                  & -                   &-           &-  \\
MSPN\cite{MSPN}            & MSPN             &$256\times192$    &-&75.9     & -            & -                   & -                  &-                &-   \\
\hline
SimpleBaseline\cite{SBNet} & ResNet-50        &$256\times192$    &23.0&71.3              & 89.9     & 78.9      & 68.3     &77.4      &76.9\\
\textbf{+UDP}              & ResNet-50        &$256\times192$    &23.0&\textbf{72.9(+1.6)}  &\textbf{90.0}    &\textbf{80.2}    &\textbf{69.7}    &\textbf{79.3}   &\textbf{78.2}\\
SimpleBaseline\cite{SBNet} & ResNet-152       &$256\times192$    &11.5&72.9             & 90.6      & 80.8     & 69.9     &79.0      &78.3\\
\textbf{+UDP}              & ResNet-152       &$256\times192$    &11.5&\textbf{74.3(+1.4)}  &\textbf{90.9}     &\textbf{81.6}    &\textbf{71.2}   &\textbf{80.6}   &\textbf{79.6}\\
SimpleBaseline\cite{SBNet} & ResNet-50        &$384\times288$    &20.3&73.2             & 90.7      & 79.9      & 69.4     &80.1     &78.2\\
\textbf{+UDP}              & ResNet-50        &$384\times288$    &20.3&\textbf{74.0(+0.8)}  &\textbf{90.3}   &\textbf{ 80.0}    &\textbf{ 70.2}  &\textbf{81.0}   &\textbf{79.0}\\
SimpleBaseline\cite{SBNet} & ResNet-152       &$384\times288$    &11.1&75.3             & 91.0      & 82.3      & 71.9    &82.0      &80.4\\
\textbf{+UDP}              & ResNet-152       &$384\times288$    &11.1&\textbf{76.2(+0.9)}  &\textbf{90.8}    &\textbf{ 83.0}   &\textbf{ 72.8}  &\textbf{82.9}    &\textbf{81.2}\\
\hline
HRNet\cite{HRNet}          & HRNet-W32        &$256\times192$    &6.9&75.6             & 91.9         & 83.0       & 72.2     &81.6     &80.5\\
\textbf{+UDP}              & HRNet-W32        &$256\times192$    &6.9&\textbf{76.8(+1.2)} & \textbf{91.9}       &\textbf{83.7}     & \textbf{73.1}   &\textbf{83.3}   &\textbf{81.6}\\
HRNet\cite{HRNet}          & HRNet-W48        &$256\times192$    &6.3&75.9               & 91.9        & 83.5       & 72.6     &82.1      &80.9\\
\textbf{+UDP }             & HRNet-W48        &$256\times192$    &6.3&\textbf{77.2(+1.3)}  &\textbf{ 91.8}      & \textbf{83.7}    &\textbf{73.8}    &\textbf{83.7}    &\textbf{82.0}\\
HRNet\cite{HRNet}          & HRNet-W32        &$384\times288$    &6.2&76.7             & 91.9       & 83.6       & 73.2     &83.2      &81.6\\
\textbf{+UDP}              & HRNet-W32        &$384\times288$    &6.2&\textbf{77.8(+1.1)} & \textbf{91.7}    & \textbf{84.5}     & \textbf{74.2}   &\textbf{84.3}   &\textbf{82.4}\\
HRNet\cite{HRNet}          & HRNet-W48        &$384\times288$    &5.3&77.1              & 91.8      & 83.8       & 73.5      &83.5     &81.8\\
\textbf{+UDP }             & HRNet-W48        &$384\times288$    &5.3&\textbf{77.8(+0.7)} & \textbf{92.0}     &\textbf{84.3}      &\textbf{74.2}    &\textbf{84.5}   &\textbf{82.5}\\
\hline

\hline
\end{tabular}
\end{center}

\end{table*}

\subsubsection{Join Analysis of Biased Coordinate System Transformation and Biased Keypoint Format Transformation.}

Error ${^o}e(x)' = \frac{1}{2s}$ in Equation~\ref{eq:errorshift1pixel} has an impact on the decoding result distribution. With a specific stride factor $s=4$ and considering Equation~\ref{eq:inference4}, we have:
\begin{equation}
    \begin{split}
    \begin{bmatrix} 1 & 0 & 1 \\
                    0 & 1 & 0 \\
                    0 & 0 & 1 \end{bmatrix}\hat{\textbf{k}}_o' &= \begin{bmatrix} 1 & 0 & 1 \\
                                                                                  0 & 1 & 0 \\
                                                                                  0 & 0 & 1 \end{bmatrix}\begin{bmatrix} 1 & 0 & \frac{1-s}{s} \\
                                                                                                                         0 & 1 & 0 \\
                                                                                                                         0 & 0 & 1 \end{bmatrix}\hat{\textbf{k}}_o\\
                    &=\begin{bmatrix} 1 & 0 & \frac{1}{s} \\
                                      0 & 1 & 0 \\
                                      0 & 0 &1 \end{bmatrix}\hat{\textbf{k}}_o\\
                    &=\begin{bmatrix} 1 & 0 & 0.25 \\
                                      0 & 1 & 0 \\
                                      0 & 0 & 1 \end{bmatrix}\hat{\textbf{k}}_o
    \end{split}
\end{equation}
As a result, the predicted heatmap from flipped image in $O_o\text{-}X_oY_o$ is changed into $\hat{\mathcal{C}}_o' = \mathcal{C}(x,y,m+0.25,n) $, and the average heatmap distribution is changed into:
\begin{equation}
\label{eq:approximation}
    \begin{split}
    \hat{\mathcal{C}}_{o,avg}   &= \frac{\mathcal{C}(x,y,m+0.25,n) + \mathcal{C}(x,y,m,n)}{2}\\
                                 &\approx \mathcal{C}(x,y,m+0.125,n)
    \end{split}
\end{equation}
where we use a approximation to simplified the following analysis. Finally, error ${^o}e(x)' = \frac{1}{2s}$ leads to a variation of the result distribution in Equation~\ref{eq:biasdecoding}:
\begin{equation}
    \hat{m}= \begin{cases}
		Floor(m) + 0.25  \!\!  & \textbf{if} \ m-Floor(m)<0.375\\
		Ceil(m) - 0.25  \!\!  & \textbf{if} \ 0.375\leq m-Floor(m)<0.875 \\
        Ceil(m) + 0.25  \!\!  & \textbf{otherwise}
	\end{cases}
\end{equation}
and the expected error in $O_o\text{-}X_o$ direction is enlarged by just $1/32$ unit length to $E(|m  - \hat{m}|) = 5/32 \approx 0.156$ with a larger variance of $V(|m  - \hat{m}|) = 37/3072 \approx 0.012$. 

Considering the error ${^o}e(x) = \frac{s-1}{2s}$ in Equation~\ref{eq:errornoshift}, the distribution of decoding result in Equation~\ref{eq:biasdecoding} will change into:
\begin{equation}
    \hat{m}= \begin{cases}
		Floor(m) - 0.25  \!\!  & \textbf{if} \ m-Floor(m)<0.375\\
		Floor(m) + 0.25  \!\!  & \textbf{if} \ 0.375\leq m-Floor(m)<0.875 \\
        Ceil(m) - 0.25  \!\!  & \textbf{otherwise}
	\end{cases}
\end{equation}
and the expected error in $O_o\text{-}X_o$ direction is enlarged by $1/4$ unit length to $E(|m  - \hat{m}|) = 3/8 = 0.375$ with a larger variance of $V(|m  - \hat{m}|) = 1/48 \approx 0.0208$. Compared with ${^o}e(x) = \frac{s-1}{2s} = 0.375$, the biased decoding method contributes a variance which will have extra negative impact on the final performance. It is worth noting that, the actual errors are more complicated than that analyzed above, as the approximation in Equation~\ref{eq:approximation} also has an impact on the errors.

\begin{table*}
\caption{The improvement of AP on COCO \textit{test-dev} set when the proposed UDP is applied to state-of-the-art methods. $\dagger$ means unreported results in the original paper and trained with official implementation by us.}
\label{tab:test-dev}
\begin{center}
\begin{tabular}{l|l|c|lcccccc}

\hline
Method                             & Backbone         &Input size  &AP   & $\text{AP}^{50}$ & $\text{AP}^{75}$ & $\text{AP}^{\text{M}}$ &$\text{AP}^{\text{L}}$ &AR  \\
\hline
\multicolumn{9}{c}{Bottom-up methods}\\
\hline

AE \cite{AssociativeEmbedding}     & Hourglass\cite{Hourglass}   &$512\times512$    &56.6                  & 81.8              & 61.8         & 49.8          &67.0           &-   \\
G-RMI\cite{G-RMI}                  & ResNet-101     &$353\times 257$  &64.9                  & 85.5              & 71.3         & 62.3          &70.0           &69.7\\

PersonLab \cite{PersonLab}         & ResNet-152    &$1401\times1401$  &66.5                  & 88.0              & 72.6         & 62.4          &72.3           &-\\
PifPaf \cite{PIFPAF}               & -             &-                 &66.7                  & -                 & -            &-              &-              &-    \\
HigherHRNet \cite{Higher}          & HRNet-W32     &$512\times512$    &64.1                  & 86.3              & 70.4         & 57.4          &73.9           &-\\
\textbf{+UDP}                      & HRNet-W32     &$512\times512$    &\textbf{66.8 (+2.7)}         & \textbf{88.2}     & \textbf{73.0}& \textbf{61.1} &\textbf{75.0}  &\textbf{71.5}\\
HigherHRNet \cite{Higher}          &HigherHRNet-W32&$512\times512$    &66.4                  & 87.5              & 72.8         & 61.2          &74.2           &-\\
\textbf{+UDP}                      &HigherHRNet-W32&$512\times512$    &\textbf{67.2 (+0.8)}         & \textbf{88.1}     & \textbf{73.6}& \textbf{62.0} &\textbf{74.3}  &\textbf{72.0}\\
HigherHRNet \cite{Higher}$\dagger$ &HRNet-W48      &$640\times640$    &67.4                  & 88.6              & 74.2         & 62.6          &74.3           &72.8\\
\textbf{+UDP}                      &HRNet-W48      &$640\times640$    &\textbf{68.1 (+0.2)}         & \textbf{88.3}     & \textbf{74.6}& \textbf{63.9} &\textbf{74.1}  &\textbf{73.1}\\
HigherHRNet \cite{Higher}          &HigherHRNet-W48&$640\times640$    &68.4                  & 88.2              & 75.1         & 64.4          &74.2           &-\\
\textbf{+UDP}                      &HigherHRNet-W48&$640\times640$    &\textbf{68.6 (+0.2)}         & \textbf{88.2}     & \textbf{75.5}& \textbf{65.0} &\textbf{74.0}  &\textbf{73.5}\\
\hline
\multicolumn{9}{c}{Bottom-up methods with multi-scale ([$\times$2,$\times$1,$\times$0.5]) test as in HigherHRNet \cite{Higher} }\\
\hline
\textbf{UDP}                     &HRNet-W32      &$512\times512$    &\textbf{69.3}         & \textbf{89.2}     & \textbf{76.0}  & \textbf{64.8} &\textbf{76.0}  &\textbf{74.1}\\
HigherHRNet \cite{Higher}$\dagger$&HRNet-W32     &$512\times512$    &68.8                  & 88.8              & 75.7           & 64.4          &75.0           &73.5\\
\textbf{UDP}                     &HigherHRNet-W32&$512\times512$    &\textbf{69.1}         & \textbf{89.1}     & \textbf{75.8}  & \textbf{64.4} &\textbf{75.5}  &\textbf{73.8}\\
HigherHRNet \cite{Higher}$\dagger$&HRNet-W48     &$640\times640$    &70.4                  & 89.7              & 77.4           & 66.4          &75.7           &75.2\\
\textbf{+UDP}                    &HRNet-W48      &$640\times640$    &\textbf{70.3}         & \textbf{90.1}     & \textbf{76.7}  & \textbf{66.6} &\textbf{75.3}  &\textbf{75.1}\\
HigherHRNet \cite{Higher}        &HigherHRNet-W48&$640\times640$    &70.5                  & 89.3              & 77.2           & 66.6          &75.8           &-\\
\textbf{+UDP}                    &HigherHRNet-W48&$640\times640$    &\textbf{70.5}         & \textbf{89.4}     & \textbf{77.0}  & \textbf{66.8} &\textbf{75.4}  &\textbf{75.1}\\
\hline
\multicolumn{9}{c}{Top-down methods}\\

\hline
Mask-RCNN\cite{Mask-RCNN}          & ResNet-50-FPN\cite{FPN}    &-          &63.1 & 87.3             & 68.7             & 57.8                   &71.4                   &-   \\
Integral Pose Regression\cite{IPR} & ResNet-101\cite{Resnet}&$256\times 256$&67.8 & 88.2             & 74.8             & 63.9                   &74.0                   &-   \\
SCN\cite{SCN}                      & Hourglass\cite{Hourglass}  &-          &70.5 & 88.0             & 76.9             & 66.0                   &77.0                   &-   \\
CPN\cite{CPN}                      & ResNet-Inception &$384\times288$       &72.1 & 91.4             & 80.0             & 68.7                   &77.2                   &78.5\\
RMPE\cite{RMPE}                    & PyraNet\cite{PyraNet}&$320\times 256$  &72.3 & 89.2             & 79.1             & 68.0                   &78.6                   &-   \\
CFN\cite{CFN}                      & -                &-                    &72.6 & 86.1             & 69.7             & 78.3                   &64.1                   &-   \\
CPN(ensemble)\cite{CPN}            & ResNet-Inception &$384\times 288$      &73.0 & 91.7             & 80.9             & 69.5                   &78.1                   &79.0\\
Posefix\cite{Posefix}              & ResNet-152       &$384\times288$       &73.6 & 90.8             & 81.0             & 70.3                   &79.8                   &79.0\\
CSANet\cite{CSANet}                & ResNet-152       &$384\times288$       &74.5 & 91.7             & 82.1             & 71.2                   &80.2                   &80.7\\
MSPN\cite{MSPN}                    & MSPN\cite{MSPN}  &$384\times288$       &76.1 & 93.4             & 83.8             & 72.3                   &81.5                   &81.6\\
\hline
SimpleBaseline\cite{CPN}           & ResNet-50     &$256\times192$       &70.2      & 90.9             & 78.3             & 67.1            &75.9            &75.8\\
\textbf{+UDP}                                         & ResNet-50     &$256\times192$       &\textbf{71.7 (+1.5)}    &\textbf{91.1}           &\textbf{79.6}           &\textbf{68.6}         &\textbf{77.5}          &\textbf{77.2}\\
SimpleBaseline\cite{CPN}           & ResNet-50     &$384\times288$       &71.3      & 91.0             & 78.5             & 67.3            &77.9            &76.6\\
\textbf{+UDP}                                         & ResNet-50     &$384\times288$       &\textbf{72.5 (+1.2)}    &\textbf{91.1}           &\textbf{79.7}           &\textbf{68.8}         &\textbf{79.1}          &\textbf{77.9}\\
SimpleBaseline\cite{CPN}           & ResNet-152   &$256\times192$       &71.9       & 91.4            & 80.1             & 68.9            &77.4            &77.5\\
\textbf{+UDP}                                         & ResNet-152  &$256\times192$       &\textbf{72.9 (+1.0)}    &\textbf{91.6}           &\textbf{80.9}           &\textbf{70.0}         &\textbf{78.5}          &\textbf{78.4}\\
SimpleBaseline\cite{CPN}           & ResNet-152   &$384\times288$       &73.8       & 91.7            & 81.2             & 70.3            &80.0            &79.1\\
\textbf{+UDP}                                         & ResNet-152   &$384\times288$       &\textbf{74.7 (+0.9)}    &\textbf{91.8}           &\textbf{82.1}           &\textbf{71.5}         &\textbf{80.8}          &\textbf{80.0}\\
\hline
HRNet\cite{HRNet}                  & HRNet-W32        &$256\times192$       &73.5     & 92.2          & 82.0       & 70.4        &79.0         &79.0\\
\textbf{+UDP}                                     & HRNet-W32        &$256\times192$       &\textbf{75.2 (+1.7)}   & \textbf{92.4}        &\textbf{82.9}     & \textbf{72.0}      &\textbf{80.8}      &\textbf{80.4}\\
HRNet\cite{HRNet}                  & HRNet-W32        &$384\times288$     &74.9     & 92.5          & 82.8       & 71.3        &80.9         &80.1\\
\textbf{+UDP}                                     & HRNet-W32        &$384\times288$       &\textbf{76.1 (+1.2)}   &\textbf{92.5}        & \textbf{83.5}    & \textbf{72.8}      &\textbf{82.0}      &\textbf{81.3}\\
HRNet\cite{HRNet}                 & HRNet-W48        &$256\times192$       &74.3     & 92.4          & 82.6       & 71.2        &79.6         &79.7\\
\textbf{+UDP}                                     & HRNet-W48        &$256\times192$       &\textbf{75.7 (+1.4)}   & \textbf{92.4}        & \textbf{83.3}    &\textbf{ 72.5}      &\textbf{81.4}      &\textbf{80.9}\\
HRNet\cite{HRNet}                  & HRNet-W48        &$384\times288$       &75.5      & 92.5          & 83.3      & 71.9        &81.5         &80.5\\
\textbf{+UDP}                                     & HRNet-W48        &$384\times288$      &\textbf{76.5 (+1.0)}   & \textbf{92.7}        & \textbf{84.0}    & \textbf{73.0}     &\textbf{82.4}      &\textbf{81.6}\\

\hline
\end{tabular}
\end{center}
\end{table*}

\section{Experiments}
\label{sec:EP}

\subsection{Result on COCO dataset}

\subsubsection{Implementation Details}
For \textit{top-down} paradigm, we take SimpleBaseline \cite{SBNet} and HRNet \cite{HRNet} as baseline and use the official implementation \footnote{https://github.com/leoxiaobin/deep-high-resolution-net.pytorch}. All training settings are preserved except for the data processing pipeline proposed in this paper. Unbiased keypoint format transformation in combined classification and regression format is used in this paradigm as default with hyper-parameters $r=0.0625 * w_o^p$ in Equation~\ref{eq:combinedencoding}. Classification format is verified in ablation study with hyper-parameters $\delta=2.0$ in Equation~\ref{eq:encodinggaussian}. During inference, HTC \cite{HTC} detector is used to detect human instances. With multi-scale test, the 80-class and person AP on COCO \textit{val} set \cite{COCO} are $52.9$ and $65.1$, respectively. The results of HRNet \cite{HRNet} and SimpleBaseline \cite{SBNet} on COCO \textit{val} set with this human detection are reproduced for fair comparison. The inference speed is tested on \textit{val} set and measured in Person Per Second (PPS). The hardware environment mainly includes a single RTX 2080ti GPU and an Intel(R) Xeon(R) E5-2630-v4@2.20GHz CPU.

For \textit{bottom-up} paradigm, we take HigherHRNet\footnote{https://github.com/HRNet/HigherHRNet-Human-Pose-Estimation} \cite{Higher} as baseline and both HRNet and HigherHRNet network structures are exploited. All training settings are preserved except for the data processing pipeline proposed in this paper. During inference, the operation of resizing the network output is removed and the decoding method is replaced with the unbiased one in Equation~\ref{eq:unbiaseddecodinggaussian}. Testing with single scale and multi-scale (i.e., [$\times$2,$\times$1,$\times$0.5], where $\times$2 means that the input resolution is enlarged by factor 2 like 512$\times$512 to 1024$\times$1024) are reported respectively. The inference speed is measured in Image Per Second (IPS).

\subsubsection{Results of top-down paradigm on the \textit{val} set.}
We report the performance improvement when UDP is applied to SimpleBaseline \cite{SBNet} and HRNet \cite{HRNet} in Table~\ref{tab:val}.
Considering the series of SimpleBaseline, the promotions are +1.6 AP (71.3 to 72.9) for ResNet-50 backbone and +1.4 AP (72.9 to 74.3) for ResNet-152 backbone. For higher network input resolution, the promotions are +0.8 AP and +0.9 AP respectively.
For HRNet family, the promotion is +1.2 AP (75.6 to 76.8) for HRNet-w32 backbone and +1.3 AP (75.9 to 77.2) for HRNet-w48 backbone. For higher network input resolution, the promotions are +1.1 AP and +0.7 AP respectively.
We summarize some key characteristics of the results: i) improvements are consistent among different backbone types, which indicates that the learning ability of the network has little impact on the precision loss caused by the biased data processing pipeline. This indicates that more powerful network structures proposed in future work would not help solving the bias problem and UDP is the necessary solution. ii) improvements on methods with smaller network input resolution are more than that with larger network input resolution. This is in line with the analysis in methodology that larger network input size can help suppressing the error and models with smaller network input size suffer more precision degression. iii) No extra latency is involved in the proposed method, which means that UDP provides the aforementioned improvement at no cost.

\subsubsection{Results of bottom-up paradigm on the \textit{val} set.}
We take the most recent method HigherHRNet \cite{Higher} as the representative baseline with two network constructions HRNet and HigherHRNet. With biased data processing as reported in \cite{Higher}, HRNet-W32-512$\times$512 configuration only scores 64.4 AP with an inference speed of 0.8 IPS and HigherHRNet-W32-512$\times$512 configuration 67.1 AP with an inference speed of 1.1 IPS. By contrast with UDP, HRNet-W32-512$\times$512 configuration scores 67.0 AP with an inference speed of 4.9 IPS which has 2.6 AP superiority and 6.1 times faster than the baseline. The performance of this configuration is even close to the baseline with HigherHRNet-W32-512$\times$512 configuration, and still 4.5 times faster than it. HigherHRNet-W32-512$\times$512-UDP configuration scores 67.8 AP with an inference speed of 2.9 IPS, which has 0.7 AP superiority and 2.6 times faster than the baseline configuration HigherHRNet-W32-512$\times$512. At no cost, UDP offers both performance boosting and latency reducing. With UDP, we have a more reasonable performance difference between HRNet-W32-512$\times$512 and HigherHRNet-W32-512$\times$512 on COCO \textit{val} set, which is $+0.8$ AP improvement at the cost of $+70\%$ extra latency in inference.

\subsubsection{Results on the \textit{test-dev} set.} Table~\ref{tab:test-dev} and Figure~\ref{fig:mAP-gflops} report the performance of UDP on COCO \textit{test-dev} set. The results show similar improvement compared with \textit{val} set, indicating the steady generalization property of UDP. Specifically, our approach promotes SimpleBaseline by 1.5 AP (70.2 to 71.7) and 1.0 AP (71.9 to 72.9) within ResNet50-256$\times$192 and ResNet152-256$\times$192 configurations, respectively. For HRNet within W32-256$\times$192 and W48-256$\times$192 configurations, UDP obtains gains by 1.7 AP (73.5 to 75.2) and 1.4 AP (74.3 to 75.7), respectively. The HRNet-W48-384$\times$288 equipped with UDP achieves 76.5 AP and sets a new state-of-the-art for human pose estimation.

\subsection{Results on CrowdPose dataset}
We utilize the CrowdPose \cite{Crowdpose} dataset to verify the generalization ability of UDP among different data distributions. HigherHRNet \cite{Higher} is used as baseline and the experimental configurations are set the same as those in COCO dataset. In line with \cite{Higher}, models are trained on \textit{train} and \textit{val} sets and tested on \textit{test} set. We report the improvement of AP on Table~\ref{tab:crowdpose}. According to the experimental results, UDP not only promotes the accuracy of all configurations, but also speeds up the inference by a large margin. This is in line with that in COCO dataset. The exceptional thing is that, when UDP is applied, HigherHRNet-W32-512$\times$512 configuration (65.6 AP with 2.4 IPS inference speed) and HigherHRNet-W48-640$\times$640 configuration (66.7 AP with 1.8 IPS inference speed) with higher output resolution doesn't show any superiority on HRNet-W32 configuration (66.1 AP with 4.5 IPS inference speed) and HRNet-W48-640$\times$640 configuration (67.2 AP with 4.2 IPS inference speed). This puts doubt on the generalization of the techniques proposed in HigherHRNet \cite{Higher}. Thus we empirically argue that, by effecting the performance and misguided the researchers, the biased data processing pipeline has a negative effect on the technology development.

\subsection{Ablation Study on Top-down Paradigm}
In this subsection, we use HRNet-W32 backbone and $256\times192$ input size to perform ablation study on the techniques involved in the data processing pipeline. Techniques we study here includes Unbiased Coordinate System Transformation (UCST), Flipping Testing (FT), Shift the Network Output by One Pixel (SNOOP) used in some state-of-the-arts\cite{SBNet,HRNet,DARK}, Extra Compensation (EC) proposed in Section~\ref{sec:decst} for the residual error left by using SNOOP, Unbiased Keypoint Format Transformation in Combined Classification and Regression Form (UKFT-CCRF), Unbiased Keypoint Format Transformation in Classification Form (UKFT-CF). Experimental settings and the corresponding performance on COCO \textit{val} set are listed in Table~\ref{tab:ablation_top_down}.

When FT is absent, configuration A and B have similar performances (74.5 AP and 74.4 AP) which is guaranteed by the establish of Equation~\ref{eq:inference1} and Equation~\ref{eq:inference3}. This verify the conjecture that using resolution counted in pixels instead of size measured in unit length when performing resizing transformation has no impact on the unbias property of the data processing pipeline. However, when FT is adopted, the performance of configuration C doesn't shows any improvement on configuration A, and instead, even drops by 1.2 AP from 74.5 AP to 73.3 AP. This showcase the tremendous negative effect of the error $e(x)_o$ reported in Equation~\ref{eq:errornoshift}. The trap caused by biased coordinate system transformation pipeline is so deep that producing great demand for remedies. By contrast with the proposed UCST, configuration D (75.7 AP) has 1.3 AP improvement on configuration B (74.4 AP). UCST is the prerequisite for performance improving with FT.

\begin{table*}[t]
\caption{The improvement of AP on CrowdPose \textit{test} set when UDP is applied. $\dagger$ means unreported results in the original paper and trained with official implementation by us.}
\label{tab:crowdpose}
\footnotesize
\begin{center}
\begin{tabular}{l|l|c|l|lcccccc}

\hline
Method                            & Backbone      &Input size             &IPS   &AP                     & $\text{AP}^{50}$ & $\text{AP}^{75}$   & $\text{AP}^{E}$   & $\text{AR}^{M}$       &$\text{AR}^{H}$ \\
\hline
SPPE \cite{Crowdpose}             & ResNet-101    &$320\times240$         &-     &66.0                   & 84.2             & 71.5               & 75.5              &66.3                   &57.4\\
HigherHRNet \cite{Higher}$\dagger$&HRNet-W32      &$512\times512$         &0.4   &65.0                   & 85.9             & 69.7               & 72.6              &65.4                   &57.7\\
\textbf{+UDP}                     &HRNet-W32      &$512\times512$         &\textbf{4.5 ($\times$11.3)}   &\textbf{66.1 (+1.1)}          & \textbf{86.7}    & \textbf{70.9}      & \textbf{73.5}     &\textbf{66.6}          &\textbf{58.2}\\
HigherHRNet \cite{Higher}$\dagger$&HigherHRNet-W32&$512\times512$         &0.7   &65.5                   & 85.9             & 70.5               & 72.8              &66.0                   &57.7\\
\textbf{+UDP}                     &HigherHRNet-W32&$512\times512$         &\textbf{2.4 ($\times$3.4)}   &\textbf{65.6 (+0.1)}          & \textbf{86.5}    & \textbf{70.5}      & \textbf{73.1}     &\textbf{66.2}          &\textbf{57.5}\\
HigherHRNet \cite{Higher}$\dagger$&HRNet-W48      &$640\times640$         &0.34   &67.0                   & 87.2             & 71.9               & 73.8              &67.7                   &59.6\\
\textbf{+UDP}                     &HRNet-W48      &$640\times640$         &\textbf{4.2 ($\times$12.4)}   &\textbf{67.2 (+0.2)}   & \textbf{87.4}    & \textbf{72.1}      & \textbf{74.5}     &\textbf{67.8}          &\textbf{59.3}\\
HigherHRNet \cite{Higher}         &HigherHRNet-W48&$640\times640$         &0.5   &65.9                   & 86.4             & 70.6               & 73.3              &66.5                   &57.9\\
\textbf{+UDP}                     &HigherHRNet-W48&$640\times640$         &\textbf{1.8 ($\times$3.6)}   &\textbf{66.7 (+0.8)}   & \textbf{86.6}    & \textbf{71.7}      & \textbf{74.2}     &\textbf{67.3}          &\textbf{59.1}\\
\hline
\multicolumn{9}{c}{Bottom-up methods with multi-scale ([$\times$2,$\times$1,$\times$0.5]) test as in HigherHRNet \cite{Higher}}\\
\hline
HigherHRNet \cite{Higher}$\dagger$&HRNet-W32&$512\times512$               &-     &67.4                   & 87.1             & 72.3               & 76.1              &67.9                   &58.6\\
\textbf{+UDP}                     &HRNet-W32&$512\times512$               &-     &\textbf{67.8 (+0.4)}          & \textbf{88.0}    & \textbf{72.7}      & \textbf{76.4}     &\textbf{68.3}          &\textbf{59.3}\\
HigherHRNet \cite{Higher}$\dagger$&HigherHRNet-W32&$512\times512$         &-     &61.4                   & 80.1             & 65.7               & 69.9              &62.7                   &50.1\\
\textbf{+UDP}                     &HigherHRNet-W32&$512\times512$         &-     &\textbf{67.5 (+6.1)}          & \textbf{87.5}    & \textbf{72.5}      & \textbf{76.1}     &\textbf{68.0}          &\textbf{58.8}\\
HigherHRNet \cite{Higher}$\dagger$&HRNet-W48      &$640\times640$         &-     &68.8                   & 88.3             & 73.9               & 76.5              &69.5                   &60.2\\
\textbf{+UDP}                     &HRNet-W48      &$640\times640$         &-     &\textbf{69.0 (+0.2)}   & \textbf{88.5}    & \textbf{74.0}      & \textbf{76.9}     &\textbf{69.5}          &\textbf{60.7}\\
HigherHRNet \cite{Higher}         &HigherHRNet-W48&$640\times640$         &-     &67.6                   & 87.4             & 72.6               & 75.8              &68.1                   &58.9\\
\textbf{+UDP}                     &HigherHRNet-W48&$640\times640$         &-     &\textbf{68.2 (+0.6)}   & \textbf{88.0}    & \textbf{72.9}      & \textbf{76.6}     &\textbf{68.7}          &\textbf{59.9}\\
\hline
\end{tabular}
\end{center}
\end{table*}

By performing an empirical compensation, configuration E with SNOOP scores 75.6 AP, which is close to the result in configuration D with UCST. This means that, by taking the unbiased configuration D as reference, $66.7\%$ of error $e(x)_o$ suppressed by SNOOP has a dominating effect on the performance, and the remainder (i.e., $e(x)_o'$) would have little impact on the performance (i.e., around 0.1 AP, 75.6$\rightarrow$75.7). We subsequently perform EC in configuration F to verify this. According to the experimental result, EC offers just 0.2 AP (75.6$\rightarrow$75.8) improvement which is in line with the aforementioned inference. We empirically blame the ineffective of EC for the insensitive of the evaluation system, where the human pose are manually annotated with a certain variance. Proving by EC, the existence of residual error $e(x)_o'$ indicates that the widely used unreported compensation (SNOOP) is a suboptimal remedy not only for its low interpretability, but also for its poorer accuracy.

\begin{table}[t]
\footnotesize
\caption{Ablation study in top-down paradigm on COCO \textit{val} set. UCST denotes Unbiased Coordinate System Transformation. SNOOP denotes Shift the Network Output by One Pixel. EC denotes Extra Compensation. UKFTCCRF denotes Unbiased Keypoint Format Transformation in Combined Classification and Regression Form, UKFTCF denotes Unbiased Keypoint Format Transformation in Classification Form.}
\begin{center}
\begin{tabular}{c|c|c|c|c|c|c|c}

\hline
ID      &FT             &UCST           &SNOOP          &EC             &UKFTCCRF       &UKFTCF        &AP   \\
\hline
A       &               &               &               &               &               &               &74.5 \\
B       &               &$\checkmark$   &               &               &               &               &74.4 \\
C       &$\checkmark$   &               &               &               &               &               &73.3 \\
D       &$\checkmark$   &$\checkmark$   &               &               &               &               &75.7 \\
E       &$\checkmark$   &               &$\checkmark$   &               &               &               &75.6 \\
F       &$\checkmark$   &               &$\checkmark$   &$\checkmark$   &               &               &75.8 \\
\hline
G       &$\checkmark$   &               &               &               &$\checkmark$   &               &74.5 \\
H       &$\checkmark$   &$\checkmark$   &               &               &$\checkmark$   &               &\textbf{76.8} \\
I       &$\checkmark$   &$\checkmark$   &               &               &               &$\checkmark$   &\textbf{76.8} \\
\hline
\end{tabular}
\end{center}
\label{tab:ablation_top_down}
\end{table}

With configuration E and I, we replace the encoding-decoding methods in configuration D with UKFT-CCRF and UKFT-CF, respectively. With UKFT, configuration E (76.8 AP) and I (76.8 AP) have similar improvement (+1.1 AP) upon baseline configuration D (75.7 AP), which indicates that the biased keypoint format transformation has a considerable impact on the performance. Beside, this also tells that the configuration (i.e., HRNet-W32 network structure with 256$\times$192 input size and training settings in \cite{HRNet}) used in this subsection has similar learning ability on the two unbiased format introduced in this paper. With configuration G where UCST is absent and only UKFT-CCRF is applied, the performance degrades by -2.3 AP to 74.5 AP. Both UCST and UKFT are important for accurate prediction and the defects in the data processing pipeline will have accumulative impact on the result.

\begin{table}[h]
\footnotesize
\caption{Ablation study of techniques in bottom-up paradigm on COCO \textit{val} set. HNOR denotes Higher Network Output Resolution, UCST denotes Unbiased Coordinate System Transformation, UKFT-CF denotes Unbiased Keypoint Format Transformation in Classification Form and RNO denotes Resize the Network Output.}
\begin{center}
\begin{tabular}{c|c|c|c|c|c|c}

\hline
ID      &HNOR           &UCST           &UKFT-CF        &RNO            &IPS            &AP    \\
\hline
A       &               &               &               &$\checkmark$   &0.8            &64.4 \\
B       &               &$\checkmark$   &               &               &4.9            &65.9 \\
C       &               &$\checkmark$   &$\checkmark$   &               &\textbf{4.9}   &\textbf{67.0} \\
D       &               &$\checkmark$   &$\checkmark$   &$\checkmark$   &0.8            &66.1 \\
\hline
E       &$\checkmark$   &               &               &               &2.9            &66.9 \\
F       &$\checkmark$   &               &               &$\checkmark$   &1.1            &67.1 \\
G       &$\checkmark$   &               &$\checkmark$   &$\checkmark$   &1.1            &67.1 \\
H       &$\checkmark$   &$\checkmark$   &               &               &2.9            &67.3 \\
I       &$\checkmark$   &$\checkmark$   &$\checkmark$   &               &\textbf{2.9}   &\textbf{67.8} \\
J       &$\checkmark$   &$\checkmark$   &$\checkmark$   &$\checkmark$   &1.1            &67.8 \\
\hline
\end{tabular}
\end{center}
\label{tab:ablation_bottom_up}
\end{table}

\subsection{Ablation Study on Bottom-up Paradigm}
In this subsection, we study how Higher Network Output Resolution (HNOR), Unbiased Coordinate System Transformation (UCST), Unbiased Keypoint Format Transformation in Classification Form (UKFT-CF) and Resize the Network Output (RNO) affect the bottom-up method HightHRNet \cite{Higher}. Flipping Testing (FT) is used as default. Experimental settings and the corresponding performance on COCO \textit{val} set are listed in Table~\ref{tab:ablation_bottom_up}.

With configuration B, We firstly remove the operation of RNO and apply UCST to the baseline configuration A (64.4 AP and 0.8 IPS). This offers a performance improvement of 1.5 AP and a speed up of 5.9 times to 65.9 AP with 4.9 IPS inference speed. By additionally applying UKFT-CF in configuration B, configuration C scores 67.0 AP with the same inference speed. Both UCST and UKFT are effective as in top-down paradigm.

The referenced configuration F with 67.1 AP and 1.1 IPS inference speed is the recommended settings in HigherHRNet \cite{Higher}. By constructing configuration E, we remove RNO from it to test the effect of this operation. And according to the result, RNO provides a negligible improvement of 0.2 AP at the high cost of 2.6 times latency in inference. With configuration H and I, we show that the proposed UCST and UKFT-CF incrementally promote the performance on configuration E by 0.4 AP to 67.3 AP and by additionally 0.5 AP to 67.8 AP, while the inference speed is maintained in 2.9 IPS. These improvements are relatively small when compared with that in configuration B and C. This is in line with the theory that HNOR helps suppress part of the systemic error hidden in data processing pipeline. When unbiased data processing is applying, the HigherHRNet-W32 backbone (i.e., configuration I) still has 0.8 AP superiority on HRNet-W32 (i.e., configuration C) but at the cost of extra $70\%$ latency in inference.

Finally, with configuration D and J, we test the impact of Resize the RNO on the results with UDP. The performance variances are 0 AP with 6.1 times latency and -0.9 AP with 2.6 times latency respectively. RNO is unnecessary for bottom-up paradigm when unbiased data processing is applied. The performance degradation in configuration D from C is derided from the distribution variation caused by the resizing operation. As this destroys the precondition of using UKFT-CF, where a gaussian distribution is strictly required \cite{DARK}.

\section{Conclusion}
In this paper, the common biased data processing for human pose estimation is quantitatively analysed. Interestingly, we find that the systematic errors in standard coordinate system transformation and keypoint format transformation couple together, significantly degrade the performance of human pose estimators in both top-down and bottom-up paradigms. A trap is laid for the research community and subsequently give born to many suboptimal remedies. This paper solves this problem by formulating a principled Unbiased Data Processing (UDP) strategy , which consists unbiased coordinate system transformation and unbiased keypoint format transformation. UDP not only pushes the performance boundary of human pose estimation, but also provides a reliable baseline for research community by wiping out the trap formulated in the defective data processing pipeline.


%

\ifCLASSOPTIONcaptionsoff
  \newpage
\fi



\bibliographystyle{IEEEtran}
\bibliography{egbib}

\end{document}